\documentclass[10pt,journal,compsoc]{IEEEtran}
\usepackage{cite}

\usepackage{times}
\usepackage{epsfig}
\usepackage{graphicx}
\usepackage{amsmath}
\usepackage{amssymb}
\usepackage{balance}
\usepackage{url}
\usepackage{multirow}
\usepackage[caption=false,font=normalsize,labelfont=sf,textfont=sf]{subfig}
\usepackage{booktabs}
\usepackage{xcolor}
\usepackage{mathrsfs}
\usepackage{colortbl}
\usepackage{changes}
\usepackage[pagebackref=true,breaklinks=true,letterpaper=true,colorlinks,bookmarks=false]{hyperref}
\usepackage[lined,ruled,vlined]{algorithm2e}
\usepackage{listings}

\newcommand{\etal}{\emph{et al.}}
\newcommand{\eg}{\emph{e.g.}}
\newcommand{\ie}{\emph{i.e.}}

\newcommand*{\figred}{\textcolor{red}}

\newcommand{\wrt}{\emph{w.r.t. }}

\newlength{\twosubht}
\newsavebox{\twosubbox}

\ifCLASSINFOpdf
\else
\fi
\begin{document}
%
\title{Rethinking Localization Map: Towards Accurate Object Perception with Self-Enhancement Maps}
%
%
%
%

\author{Xiaolin~Zhang,
        Yunchao~Wei,
        Yi~Yang,
        ~and~Fei~Wu
\thanks{X. Zhang, Y. Wei, Y. Yang are with the Centre for Artificial Intelligence, University of Technology Sydney, Ultimo, NSW 2007, Australia.
F. Wu is with the College of Computer Science, Zhejiang University, Hangzhou, China.
(~e-mail: Xiaolin.Zhang-3@student.uts.edu.au, Yunchao.Wei@uts.edu.au, Yi.Yang@uts.edu.au,
Wufei@cs.zju.edu.cn~)}

}

\IEEEtitleabstractindextext{%
\begin{abstract}
Recently, remarkable progress has been made in weakly supervised object localization (WSOL) to promote object localization maps. 
The common practice of evaluating these maps applies an indirect and coarse way,~\ie, obtaining tight bounding boxes which can cover high-activation regions and calculating intersection-over-union (IoU) scores between the predicted and ground-truth boxes.
This measurement can evaluate the ability of localization maps to some extent, 
but we argue that the maps should be measured directly and delicately,~\ie, comparing the maps with the ground-truth object masks pixel-wisely.
To fulfill the direct evaluation, we annotate pixel-level object masks on the ILSVRC~\cite{2009-imagenet} validation set. 
We propose to use IoU-Threshold curves for evaluating the real quality of localization maps. 

Beyond the amended evaluation metric and annotated object masks, this work also introduces a novel self-enhancement method to harvest accurate object localization maps and object boundaries with only category labels as supervision.
We propose a two-stage approach to generate the localization maps by simply comparing the similarity of point-wise features between the high-activation and the rest pixels.
Based on the predicted localization maps,  we explore to estimate object boundaries on a very large dataset.
A hard-negative suppression loss is proposed for obtaining fine boundaries.
We conduct extensive experiments on the ILSVRC and CUB~\cite{WahCUB_200_2011} benchmarks.
In particular, the proposed Self-Enhancement Maps achieve the state-of-the-art localization accuracy of 54.88\% on ILSVRC. 
The code and the annotated masks are released at \url{https://github.com/xiaomengyc/SEM}.
\end{abstract}

}

\maketitle

\IEEEdisplaynontitleabstractindextext

%
\IEEEpeerreviewmaketitle

\section{Introduction}\label{sec:introduction}
\IEEEPARstart{L}{ocalization}
maps (\textit{a.k.a} class activation maps) are profoundly studied in recently years~\cite{zhou2015cnnlocalization, zhang2018adversarial, zhang2018self}.
It is originally proposed to visualize the high-level activations of classification networks~\cite{zhou2015cnnlocalization}, where object regions are expected to have higher activation scores while background regions have lower scores.
Afterward, localization maps are found to be greatly useful in Weakly Supervised Object Localization (WSOL)~\cite{zhou2015cnnlocalization,zhang2018adversarial,zhang2018self,Choe_2019_CVPR,singh2017hide} tasks.
Generally, WSOL employs a post-processing step to infer object bounding boxes from the generated object localization maps. To be specific, the post-processing step first binarizes the localization maps with fixed thresholds and then obtains the object bounding box by drawing rectangles over the largest connected area.
To achieve superiority performance on WSOL, researchers seek methods for increasing the quality of localization maps and conducting the evaluation by calculating the accuracy of the inferred bounding boxes~\cite{zhang2018adversarial,zhang2018self,Choe_2019_CVPR,Xue_2019_ICCV}.
Therefore, the quality of the generated localization maps is actually measured via an indirect way with the post-processed bounding boxes.

\begin{figure*}[t]
  \centering
  \includegraphics[width=0.90\textwidth]{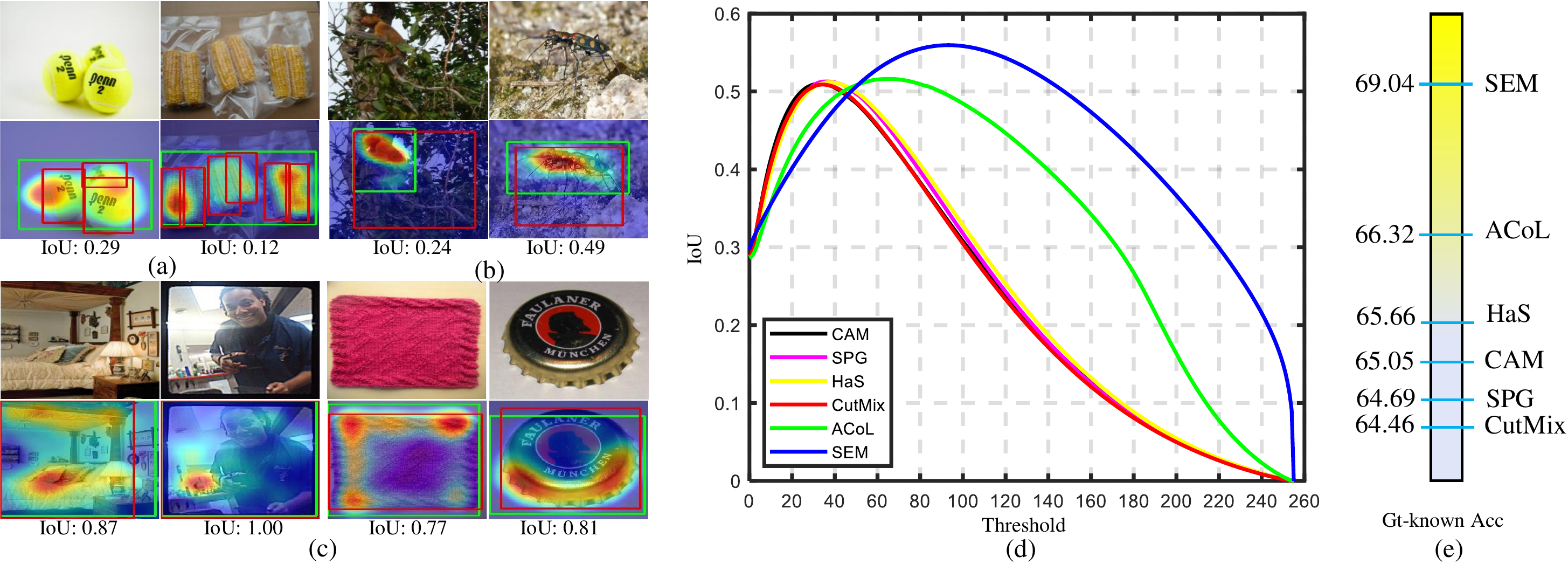}
  \vspace{-5pt}
  \caption{
  Three scenarios where the current indirect evaluation metric fails to measure localization maps: 
  (a) Localization maps accurately highlight target objects, while the metric considers them as false positive; (b) main parts of the target objects are highlighted,  while the metric gives them zero credit; (c) although the predicted bounding boxes are accurate, the localization maps fail to highlight important areas of the target objects. (Ground-truth boxes are in \textcolor{red}{red}, and the predicted are in \textcolor{green}{green}.)
  (d) The proposed IoU-Threshold curves of various approaches for direct pixel-wise measurement on our annotated test set. 
  Our Self-Enhancement Maps (SEM) achieve the best in both IoU and Threshold at the peak point.
  (e) The Gt-known localization accuracies \wrt different approaches. The proposed SEM achieves the best accuracy. 
  }\label{fig:0}
  \vspace{-17pt}
\end{figure*}

However, such an indirect measurement is not perfect for evaluating the real  accurateness of localization maps.
Figure~\ref{fig:0}\figred{a},~\ref{fig:0}\figred{b} and~\ref{fig:0}\figred{c} show three situations where the current indirect measurement fails to rightly evaluate the acquired localization maps. Ground-truth and inferred bounding boxes are shown in \textcolor{red}{red} and \textcolor{green}{green} rectangles, respectively.
First, localization maps correctly highlight the target regions,  but the indirect metric considers them as false positives, as shown in Figure~\ref{fig:0}\figred{a};
Second, the maps are successful in highlighting the most important parts of the target objects, but the localization criterion gives it zero credit as the Intersection-over-Union (IoU) is smaller than 50\% as shown in Figure~\ref{fig:0}\figred{b};
Third, although the predicted boxes accurately match the ground-truth boxes, the localization maps involve many noises or fail to highlight important areas of target objects as shown in Figure~\ref{fig:0}\figred{c}. 
(We show more such cases in the supplementary material.)
These observations reveal a significant problem that \emph{the current indirect evaluation metric is not flawless to truly reflect the correctness of the localization maps generated by WSOL algorithms.}
Therefore, we consider that it is necessary to perform a precise measurement for evaluating the quality of localization maps in a \textit{direct} way.

By annotating pixel-level object masks on the most widely applied dataset,~\ie, ILSVRC~\cite{2009-imagenet,ILSVRC15}, 
we propose a direct measurement to pixel-wisely compare the object-related partitions with our annotated ground-truth masks for evaluating the real and deliberate ability in highlighting target objects.
In particular, we calculate the object Intersection-over-Union (IoU)~\wrt different thresholds for splitting the object pixels from the background.
The highest IoU values are considered to be the best potential in localizing object regions.
Figure~\ref{fig:0}\figred{d} depicts the IoU-Threshold curves of several recently proposed methods,~\ie, CAM~\cite{zhou2015cnnlocalization}, HaS~\cite{singh2017hide}, ACoL~\cite{zhang2018adversarial}, SPG~\cite{zhang2018self} and CutMix~\cite{yun2019cutmix}.
These methods all aim at achieving better WSOL performance by promoting the quality of the produced localization maps.
For example, HaS~\cite{singh2017hide} and CutMix~\cite{yun2019cutmix} propose to hide some pitches of the input images to force classification networks to highlight more robust discriminative regions.
ACoL~\cite{zhang2018adversarial} erases the regions with the highest scores in high-level feature maps, and re-learn more object pixels by feeding the erased features into an auxiliary branch.
SPG~\cite{zhang2018self} learns a stage-wise binary salient map for each image using a side branch, and injects the pixel-level correlation into the classification network using an auxiliary loss function.
According to the proposed direct evaluation metrics,  we are surprised to find that these methods based on CAM,~\ie, HaS, CutMix, and SPG, which are claimed to improve the quality of localization maps, actually only make marginal progress. 
It turns out that \emph{the results of the indirect measurement are not consistent with the proposed direct evaluation metric}.
Especially, in the proposed pixel-wise measurement, the performance rank of these methods is ACoL $>$ HaS $>$ CAM $>$ SPG $>$ CutMix, while the claimed performance rank ~\wrt the bounding box accuracy is HaS $>$ ACoL $>$ CutMix $>$ SPG $>$ CAM, as illustrated in Figure~\ref{fig:0}\figred{d} and~\ref{fig:0}\figred{e}.
This observation supports the idea that the current indirect measurement is not perfect, and we should augment it by applying the finer pixel-wise evaluation. 

Except for the amendment in evaluating localization quality, the proposed direct measurement,~\ie, IoU-Threshold curve, can also measure the visualization effect, which is another crucial property of localization maps.
Particularly, in the IoU-Threshold curve,~\eg, Figure~\ref{fig:0}\figred{d}, the threshold~\wrt the highest IoU point can reflect the visual effect of the target object in the localization maps. 
The higher threshold values indicate the higher brightness of the object areas with the largest IoU scores, which presents better visual effects.
We notice the values of the best thresholds are rather small ($<$65 in the scale of [1,255]) with the localization maps produced by the existing methods.
Such small values can not bring enough brightness when visualizing these localization maps.

In order to overcome the two disadvantages of localization maps,~\ie, low accuracy and low brightness of object areas, we propose a simple yet effective enhancement strategy. 
Intuitively, within one object, pixels with similar appearances usually share similar features in their embedding space.
We can find the most discriminative areas and their corresponding features. 
These features can naturally be applied to capture more object regions, leading to better localization maps accordingly.
In particular, given a trained classification network, we first extract object localization maps using the CAM variant proposed in~\cite{zhang2018adversarial}. 
Then, we identify a set of discriminative seeds by ranking the scores in the obtained localization maps and extract their corresponding feature vectors from high-level feature maps. 
Finally, by performing a self-enhancement process of calculating the cosine similarity between the seed vectors and the rest, we can simply obtain the refined localization maps, named Self-Enhancement Maps (SEM). 
Surprisingly, the generated SEM can not only improve the localization accurateness in IoU (from 50.13\% to 55.67\%), but also promote the visual effect in increasing threshold of peak points from 33 to 93 after applying SEM to CAM. 
Our SEM method also significantly promote localization maps from 62.68\% to 69.04\% in the indirect evaluation method of measuring the inferred bounding boxes, surpassing the runner-up,~\ie, HaS by 2.64\%.
Besides, with the assistance of SEM, we advance the WSOL a step further by exploring to predict much finer details,~\ie, object boundaries, in a weakly supervised manner. 
Concretely, we generate the pseudo object boundaries from SEM, which are further employed as supervision to train the boundary detection models. 
As far as we know, this is the first feasible attempt to date, which can reversely validate the effectiveness of our SEM in capturing fine details of the target object. 

To sum up, we rethink the WSOL from three aspects in this work and make contributions accordingly:
\vspace{-3pt}
\begin{itemize}
    \vspace{-2pt}
    \item \emph{Is the current evaluation metrics flawless?} No. We propose a new metric,~\ie, IoU-Threshold curve, to augment the evaluation criteria of WSOL for a fairer comparison. To fulfill this, we manually annotate pixel-level masks on the ILSVRC validation set to realize the amended evaluation metrics.
    
   
    \vspace{-2pt}
    \item \emph{Can we enhance localization maps of the existing methods by a simple yet effective way?} 
    Yes. Existing methods based on CAM,~\ie, SPG, HaS, CutMix and ACoL, only make marginal progress under our precise pixel-level evaluation metric.
    Also, the thresholds \wrt the peak IoU point are small, which will result in the low brightness in visualization.
    We propose a method~\ie, Self-Enhancement Map (SEM), to enhance localization maps by employing feature similarities. 
    The proposed SEM can significantly improve both the object accuracy and brightness in localization maps.
    
    \vspace{-2pt}
    \item \emph{Can we apply localization maps to acquire more details of target objects rather than rough locations?} Yes. We explore the potential of our SEM in predicting object boundaries. To the best of our knowledge, this is the first feasible attempt to predict object boundaries using only image-level labels as supervision on a large-scale dataset.
\end{itemize}

\section{Related Work}
\vspace{-1mm}
\noindent \textbf{Weakly supervised object localization}
aims to predict object positions given only image-level labels as supervisions.
The common practice is to extract localization maps from classification networks~\cite{zhou2015cnnlocalization,zhang2016top,chattopadhay2018grad,zhu2017soft,zhou2018weakly}, 
and the localization maps can be applied as an alternative cheaper way for obtaining tight bounding boxes of target objects.
CAM~\cite{zhou2015cnnlocalization} is the most widely acknowledged method for generating class-specific localization maps.
GradCAM~\cite{selvaraju2017grad,chattopadhay2018grad} uses the gradients of target concept. 
The gradients flow into the final convolutional layers to produce a coarse localization map highlighting important regions in the image.
MWP~\cite{zhang2016top} takes a top-down strategy and probabilistic winner-take-all process to generate rough object locations.
Based on CAM, object erasing methods~\cite{singh2017hide,wei2017object,zhang2018adversarial,zhou2019dual} are proposed to cover more integral object regions by forcing the networks to learn more object patterns.
Specifically, HaS~\cite{singh2017hide} is proposed to erase some parts within the input images randomly.
CutMix~\cite{yun2019cutmix} adopts a different strategy of filling the erased area with other image patches and mixes the corresponding classification labels while training to augment the input images.
Wei~\etal~\cite{wei2017object} adopt an iterative approach to erase a small part of the most discriminative regions discovered by a trained classification network.
ACoL~\cite{zhang2018adversarial} improves the efficiency by getting the localization maps online and erasing discriminative regions within high-level feature maps.
ADL~\cite{Choe_2019_CVPR} further promotes the localization maps by applying dropout on multiple intermediate feature maps. 
Besides the methods based on the erasing operation, SPG~\cite{zhang2018self} tries to learn pixel-level correlations by applying salient masks as the auxiliary supervision to increase the quality of localization maps.
DANet~\cite{Xue_2019_ICCV} adopts a divergent activation method for learning better localization maps.

\noindent \textbf{Weakly supervised semantic segmentation} (WSSS)
aims to predict precise pixel-level object masks using weak annotations which are divided into four groups,~\ie, bounding boxes~\cite{khoreva2017simple,dai2015boxsup,2015-papandreou-weakly}, scribbles~\cite{lin2016scribblesup,tang2018normalized,tang2018regularized}, points~\cite{qian2019weakly} and image-level labels~\cite{Shimoda_2019_ICCV,Zeng_2019_ICCV,Shen_2019_CVPR,Song_2019_CVPR}.
Particularly, segmentation methods based on object bounding boxes basically apply unsupervised methods to obtain pseudo-masks and then train segmentation networks using the inferred pseudo-masks.
Methods supervised by scribble lines generally try to improve the segmentation accuracies by mining more pixels that have similar features with the annotated pixels.
For example, ScribbleSup~\cite{lin2016scribblesup} adopts an alternating solution to learn a decent segmentation network. It applies a graphical model to expand the discovered pixels and generate better pseudo-masks while uses the masks as supervision to train segmentation networks. 
Point-based methods use even weaker supervision to learn segmentation models.
PointSup~\cite{qian2019weakly} performs metric learning upon annotated pixels of the same categories across images so that objects with different content meaning can be better found.
Our WSOL shares much similarity with the segmentation methods based on using image-level labels as supervision.
Both two tasks train classification networks for getting object localization maps.
Differently, WSSS is a final task for pixel-wise classification.
Localization maps usually serve as intermediate outputs to provide some initial cues of object locations in WSSS tasks. In the main time, WSSS often applies multiple sophisticated techniques for getting better pseudo masks so that reliable segmentation models can be learned. 

\noindent \textbf{Weakly supervised object detection} (WSOD) aims at drawing tight bounding boxes of the target objects using only image-level supervision. 
Different from our WSOL task, WSOD considers a more challenging scheme where multiple objects of different semantics and scales may be distributed anywhere in the given image.
These methods generally include two aspects,~\ie, 1) estimating pseudo object bounding boxes, and 2) training the object detector using the off-the-shelf object detection approaches. 
Particularly, these methods can be roughly divided into two types according to the training manners,~\ie, alternating training~\cite{cinbis2016weakly,jie2017deep,li2016weakly,shen2018generative} and end-to-end training~\cite{bilen2016weakly,kantorov2016contextlocnet,diba2017weakly,wei2018ts2c,tang2018weakly}.
The alternating training approaches usually need first to mine positive object proposals, and then optimize detection networks towards these bounding boxes.
For instance, OM~\cite{li2016weakly} first uses a mask-out strategy to collect class-specific object proposals, and then apply multiple instances learning to mine confident candidates.
These candidates are employed as supervision to train regression networks.
FV~\cite{cinbis2016weakly} implements a multi-fold multiple instance learning procedure, which prevents alternative training from prematurely locking onto erroneous object locations.
HCP~\cite{wei2015hcp} develops a self-taught learning method to select more reliable seed positive proposals, and these proposals are further used to learn better detectors~\cite{jie2017deep}.
The alternating approach usually updates detectors and object bounding boxes alternately to lift the detection performance progressively.
For the end-to-end training approaches, they assemble the pseudo object bounding box generation and object detection into a unified framework.
For example, WSDDN~\cite{bilen2016weakly} implements two branches for classifying object categories and regressing object bounding boxes.
The outputs of the two branches are combined to judge the correctness of the predicted boxes and classes.
MELM~\cite{wan2018min} introduces a min-entropy function to measure the randomness of object locations during training, which can reduce the variance of positive instances. A recurrent learning algorithm is further applied to transfer weak supervision information to object locations progressively.
WSOD$^2$~\cite{zeng2019wsod2} proposes a joint strategy of considering bottom-up and top-down objectness to estimate the objectness scores of the regressed bounding boxes using an adaptive linear combination approach so that the correct bounding boxes and class labels can be selected.

\noindent \textbf{Object boundary detection} is a fundamental problem in understanding images.
Various edge detectors have been proposed and achieve many successes, including supervised methods, weakly supervised methods, and unsupervised methods.
Canny~\cite{canny1986computational} is the most widely used unsupervised detector because of its efficiency.
Based on the recent convolutional networks, HED~\cite{xie2015holistically} applies multiple intermediate feature maps for predicting edges.
RCF~\cite{liu2017richer} and BDCN~\cite{he2019bi} adopt similar ideas to incorporate multiscale feature maps for the detectors.
Khoreva~\etal~\cite{khoreva2016weakly} propose a weakly supervised method for detecting object boundaries.
They infer the object boundaries with the help of segmentation masks and object bounding box proposals produced by unsupervised segmentation methods and object detectors.
However, all of these methods target on small datasets.
SOBD~\cite{uijlings2015situational} proposes to predict object boundaries on a very large dataset,~\ie, ILSVRC, by learning many detectors for different situations,~\eg, predicting one edge detector for each category.
Different from the above methods, we propose a unified approach to predict the boundaries on ILSVRC with only image-level labels as the supervision.
To the best of our knowledge, this approach is the first method trying to predict object boundaries for such a large dataset to date.

\begin{figure*}[htbp]
  \centering
    \sbox\twosubbox{%
      \resizebox{\dimexpr.9\textwidth-1em}{!}{%
        \includegraphics[height=3cm]{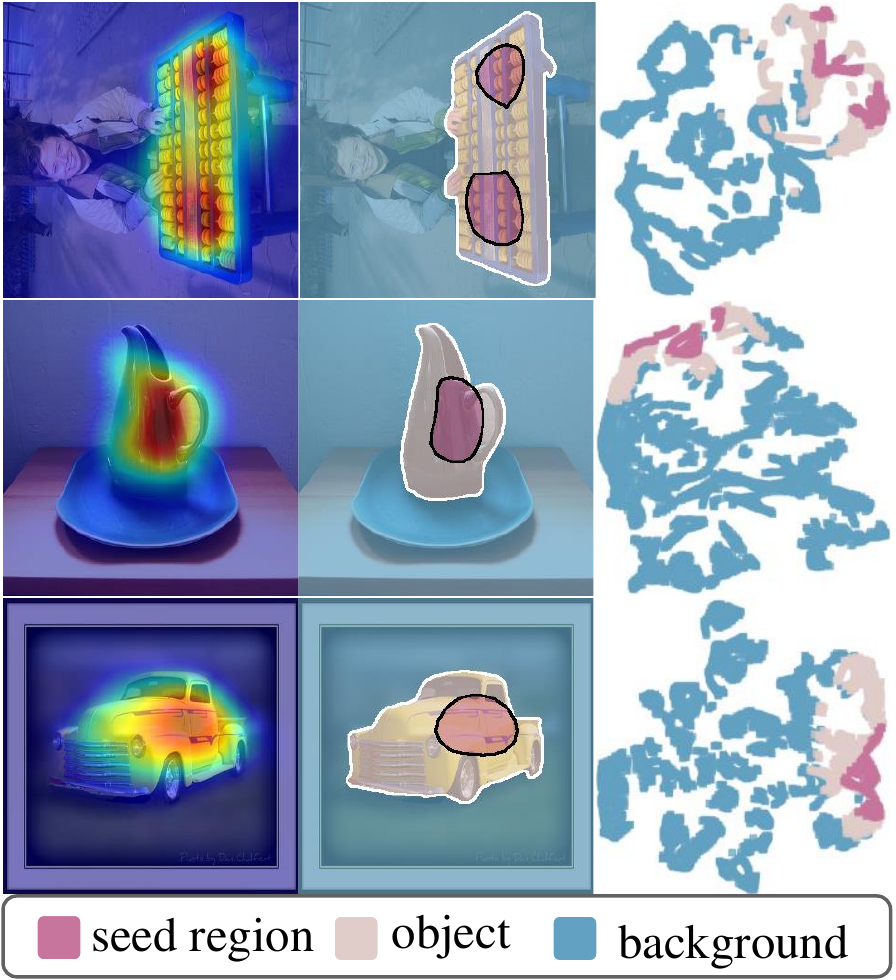}%
        \includegraphics[height=3cm]{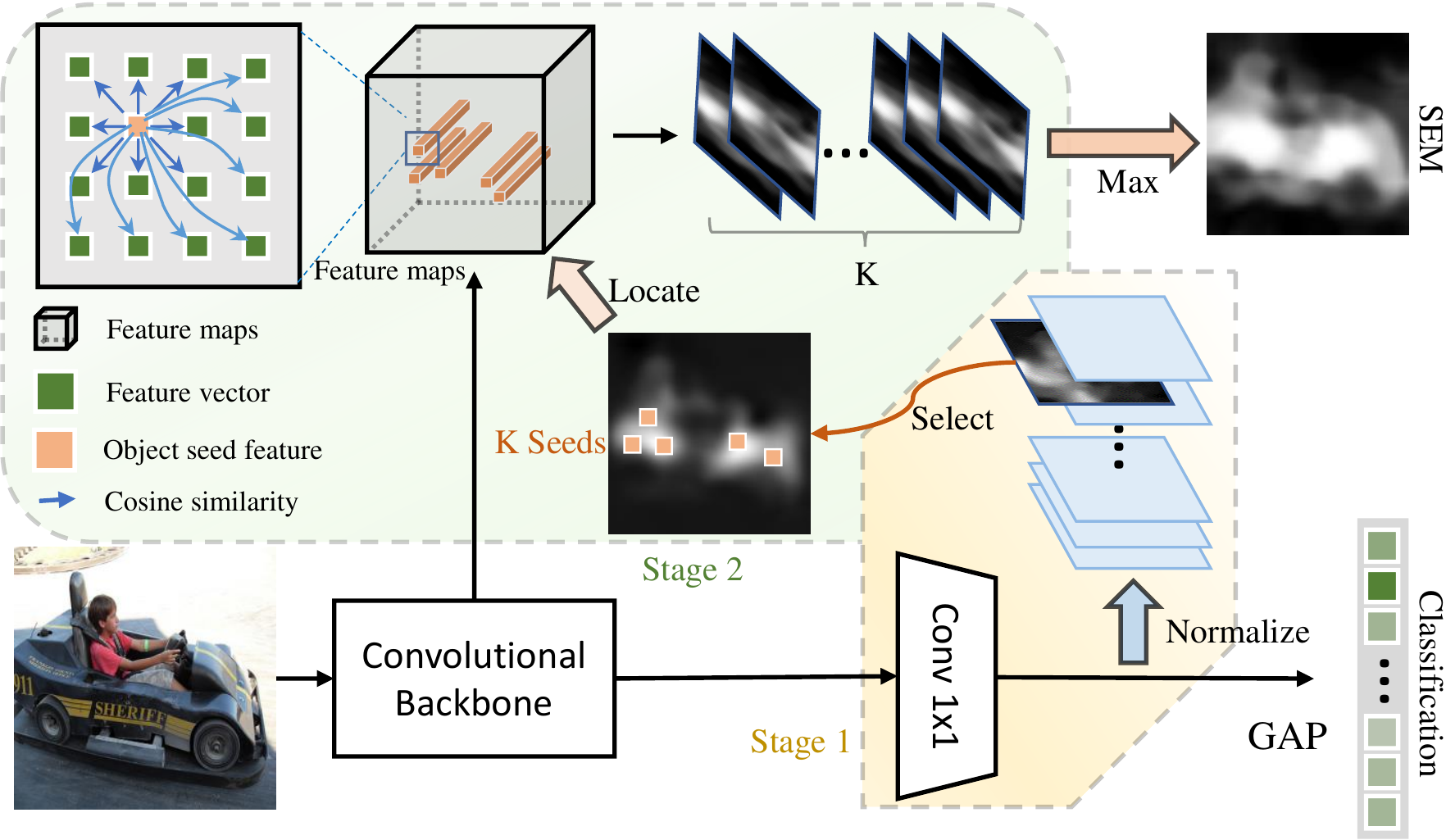}%
      }%
    }
    \setlength{\twosubht}{\ht\twosubbox}

    \centering
    \subfloat[]{\label{fig:tsne}\includegraphics[height=\twosubht]{body/figs/tsne0.pdf}}
    \qquad 
    \subfloat[]{\label{fig-framework1}\includegraphics[height=\twosubht]{body/figs/scm-framework.pdf}}
    
  \caption{(a) t-SNE~\cite{van2014accelerating} visualization of high-level features.
  Features of seed regions are close to the rest object regions while distant to the background;
 (b) The framework of SEM. In Stage 1, localization maps are extracted from the last-layer feature maps in a classification network. 
  In Stage 2, K seeds with the highest scores are firstly selected.
  Then, we calculate the similarity between the seeds and the rest pixels for getting K similarity maps.
  We finally obtain the enhanced maps by computing element-wise maximum values among the K maps.}
  \vspace{-4mm}
\end{figure*}

\section{Methodology}
In this section, we first provide the details of the new evaluation metric, which facilitates localization maps to be evaluated in a \emph{direct} manner (Section~\ref{subsec-metric}).
We then propose a simple yet effective self-enhanced approach to promote the quality of localization maps (Section~\ref{subsec-scm}). 
Benefiting from the high-quality object localization maps produced by the proposed approach, we further explore to harvest more details of detecting object boundaries in a weakly supervised manner, which is the first attempt on a large dataset (Section~\ref{subsec-edge}).

\subsection{Evaluation Metric Amendment}\label{subsec-metric}
Due to the lack of object masks, the current bounding-box-based evaluation metric alternatively adopts an \emph{indirect} way to examine the quality of localization maps.
However, as shown in Figure~\ref{fig:0}\figred{a},~\ref{fig:0}\figred{b} and~\ref{fig:0}\figred{c}, such an \emph{indirect} metric is not perfect to truly reflect the real quality of localization maps.
The concurrent work~\cite{choe2020evaluating} also notice this drawback of the indirect measurement, but \cite{choe2020evaluating} only applies precision and recall properties to reflect the quality of localization maps.
The most accurate manner for evaluating localization maps is to compare the maps to the ground-truth masks at every pixel in a \emph{direct} way.
To fulfill the purpose of the accurate and direct evaluation, we annotate object masks on the ILSVRC~\cite{2009-imagenet} validation set. 

The value of each pixel on localization maps represents the probability of belonging to target objects.
The probability scores are continuous values, and thereby, we adopt a dense binarization strategy to evaluate the IoU scores at every binarized threshold.
Particularly, we first use many equally spaced thresholds to binarize the localization maps.
Then, we calculate the IoU scores between the binarized maps and their ground-truth masks. 
To this end, a high-quality localization map should meet two requirements: 1) the entire object region can be accurately extracted with a specific threshold; 2) brightness values of pixels belonging to object and the background should differ greatly so that the objects can be well visualized. 
To present the two properties, we propose to apply IoU-Threshold curves for evaluation, as shown in Figure~\ref{fig:0}\figred{d}.
\textbf{Peak-IoU$\in [0,1]$} and \textbf{Peak-T $\in [0,255]$} denote the best IoU score and its corresponding threshold, respectively. 
If a model generates better localization maps, the peak point in the corresponding curve should drift towards the upper-right corner of the figure.
In other words, higher Peak-IoU and Peak-T values indicate better localization maps.

Based on this evaluation metric, we find that the recent efforts~\cite{zhang2018adversarial,zhang2018self,Choe_2019_CVPR,yun2019cutmix} built on CAM~\cite{zhou2015cnnlocalization} for boosting the quality of localization maps have only made marginal progress even with more sophisticate structures and extra training resources.
Another disadvantage of these methods is that the maps perform poorly in visualization.
The object regions do not have high brightness and cannot show enough distinctions between objects and background regions.
In order to increase localization ability and boost the visualization effect, we find a more promising way to enhance  these method in a training-free way, which will be introduced in the next subsection.

\vspace{-2mm}
\subsection{Self-Enhancement Map}\label{subsec-scm}
Object localization maps can highlight the object regions of interest while suppressing the values of background pixels~\cite{zhou2015cnnlocalization, zhang2016top}.
The best-known method, CAM~\cite{zhou2015cnnlocalization}, applies a category-wise fully connected layer to aggregate high-level feature maps so that class-specific localization maps can be harvested. 
These maps are typically employed as a coarse indication of the target object positions.
Nevertheless, these localization maps can only identify several small and sparse object regions while hard to highlight the entire object, which deviates from the requirement of WSOL. 
Most previous efforts~\cite{zhang2018adversarial,zhang2018self, Choe_2019_CVPR} reach a consensus that this issue is caused by the inconsistent discriminative ability of features from objects. However, even many promising solutions~\cite{zhang2018adversarial,zhang2018self, Choe_2019_CVPR}  are proposed to tackle this issue, the achievements are actually very limited based on the \emph{direct} evaluation metric shown in Figure~\ref{fig:0}\figred{d}. 

To improve the quality of localization maps, we conduct the following analysis. 
First, we employ CAM~\cite{zhou2015cnnlocalization} to produce localization maps and divide all pixels into three groups,~\ie, seed regions, the rest object regions \wrt ground-truth masks, and background regions.
The seed regions are pixels with scores greater than 0.7 (value chosen in SPG~\cite{zhang2018self}).
Then, we extract the corresponding high-level features and visualize their distributions using t-SNE~\cite{van2014accelerating}.
As shown in Figure~\ref{fig:tsne}, although some object regions do not have high activation values in the localization maps, they are still more similar to the discovered seed regions than  the background regions in the embedding feature space.
This observation motivates us that high-quality localization maps can be simply obtained by considering the pixel-wise similarity in the embedding space. To this end, we propose Self-Enhancement Map (SEM), which can truly improve the localization ability in a much simpler yet more effective way.

Concretely, the proposed SEM is a two-stage method, which can be used in conjunction with the current state-of-the-art WSOL approaches~\cite{zhou2015cnnlocalization,zhang2018self,singh2017hide,yun2019cutmix}. For simplicity, we adopt CAM~\cite{zhou2015cnnlocalization} as the first stage method of our SEM. Given a trained classification network, we extract the first-stage localization maps using the method in ACoL~\cite{zhang2018adversarial}, which is a simple variant of CAM.
Since the convolutional operations for generating the localization maps can relatively preserve positions of input pixels, the feature vectors with high activation scores are often the most discriminative and best features representing the target objects. Therefore, in the second stage, we first identify the most discriminative seeds by ranking the scores in the obtained localization maps.
Then, the corresponding feature vectors of the seeds are extracted from high-level feature maps.
We acquire similarity maps by calculating the cosine similarity between the seed vectors and the rest pixels in the high-level feature space.
To increase the robustness, we apply multiple object seeds to generate similarity maps, which are employed to produce the final self-enhanced localization maps with pixel-wise maximum aggregation. 
Despite the simplicity of our SEM, it turns out to be good at increasing both localization ability and visual effect.

Figure~\ref{fig-framework1} depicts the workflow of the proposed SEM method. 
Given an input image $I$ with its class label $y \in [0, Y-1]$,~\eg, \textit{car}, 
we forward $I$ through the classification network yielding the high-level feature maps $F$, where $Y$ is the total number of categories.
In the first stage, we adopt the method in ~\cite{zhang2018adversarial} to obtain the first-stage maps which are proved to be the same with CAM~\cite{zhang2018adversarial}.
Particularly, we apply a convolution operation as the last layer to produce class-specific feature maps which are denoted as $M' \in \mathbb{R}^{Y\times W\times H}$, where $W$ and $H$ are the width and height of the maps.
We select the $y_{th}$ feature map $M'_{y}$ from the feature maps to get the localization map of the category $y$. 
We obtain the normalized maps $M_y\in [0,1]^{W\times H}$ by following   $M_{y}(i,j)=\frac{M^{'}_{y}(i,j)-\min(M^{'}_{y})}{\max(M^{'}_{y}) - \min(M^{'}_{y})}$, where $i$ and $j$ are the indices of the map.
To distinguish the localization map produced by CAM from SEM, we denote the normalized localization map of category $y$ as $M_{cam}$ from here.
In the second stage, we choose object seed pixels whose scores in localization maps are ranked at the top K, where $K\in [1, W\times H]$ is the number of the seeds.
The corresponding feature vectors of the seeds can be extracted from the feature maps $F$ according to the positions of the seed pixels. 
For each seed vector, we calculate the cosine similarity with every pixel in $F$, getting K similarity maps $\{S_{k} | k=0,1,...,K-1\}$.
We obtain the final self-enhancement localization maps $M_{sem}$ by getting the maximum values among the K maps at each position,~\ie, $M_{sem}(i,j)=\max{(S_{k}(i,j))},k=0,1,...,K-1$.
Figure~\ref{fig-comp-box} compares the localization maps between the proposed SEM and CAM.

Figure~\ref{alg-scm} depicts the code snippet of the implementation of SEM in PyTorch~\cite{pytorch}.
We obtain the improved localization map of category $y$, only using a few lines of codes.
Line 9-10 are for the first stage and they obtain the localization map of class $y$ by selecting the $y_{th}$ feature map and normalizing the selected map.
Line 13-26 are for the second stage.
We first extract the feature vectors of the seeds,~\ie,  $F_{seeds}$, after getting the indices of pixels with top K scores.
Then, we obtain the SEM maps by computing the cosine similarity between the seed features $F_{seeds}$ and the rest features in $F$.
In our implementation, the height $H$ and width $W$ of the feature maps $M$ and $F$ are one-eighth of the input images and the number K of seeds is experimentally set as small values ($<$150), which makes the computation of cosine similarity quite efficient.

\begin{figure}
\noindent\quad
\begin{minipage}{1.0\textwidth}
\centering
\definecolor{codegreen}{rgb}{0,0.6,0}
\definecolor{codegray}{rgb}{0.5,0.5,0.5}
\definecolor{codepurple}{rgb}{0.58,0,0.82}
\definecolor{backcolour}{rgb}{0.95,0.95,0.92}
\definecolor{codeblue}{rgb}{0.25,0.5,0.5}

\lstdefinestyle{mystyle}{
    keywordstyle=\color{magenta},
    numberstyle=\tiny\color{codegray},
    stringstyle=\color{codepurple},
    basicstyle=\ttfamily\footnotesize,
    commentstyle=\color{codeblue},
    breakatwhitespace=false,         
    breaklines=true,                 
    captionpos=b,                    
    keepspaces=true,                 
    numbers=left,                    
    numbersep=5pt,                  
    showspaces=false,                
    showstringspaces=false,
    showtabs=false,                  
    tabsize=2,
    frame=lines,
    keywords={def,return},
    escapeinside={(*@}{@*)} 
}
\lstset{style=mystyle}

\begin{lstlisting}[linewidth=0.47\textwidth]
(*@\textbf{\textcolor{magenta}{def} SEM}@*)(M,F,y,K):
  (*@\textcolor{codeblue}{\#\textit{M is top-layer feature maps of size (Y,W,H)}}@*)
  (*@\textcolor{codeblue}{\#\textit{F is high-level feature maps of size (C,W,H)}}@*)
  (*@\textcolor{codeblue}{\#\textit{K is the seed number}}@*)
  (*@\textcolor{codeblue}{\#\textit{y is the category label}}@*)
  
  c, w, h = F.shape
  (*@\textcolor{codeblue}{\#Stage one}@*)
  cam = M[y, :, :]
  cam =(cam - cam.min())/(cam.max() - cam.min()) 
  
  (*@\textcolor{codeblue}{\#Stage two}@*)
  _, topk_indices = torch.topk(
                    cam.view(-1),
                    K, 
                    largest=True)
              
  F_seeds = F.view(c, -1)[:, topk_indices]
  simi = torch.nn.functional.cosine_similarity( 
                F.view(c, -1, 1), 
                F_seeds.view(c, 1, -1), 
                dim=0)
   
  sem, _ = simi.max(dim=1) 
  sem = sem.view(w, h)
  sem = (sem - sem.min())/(sem.max() - sem.min())
  
  (*@\textbf{\textcolor{magenta}{return}}@*)  sem 


\end{lstlisting}
\end{minipage}
\vspace{-10pt}
\caption{Implementation of SEM in PyTorch~\cite{pytorch}.}
\label{alg-scm}
\vspace{-10pt}
\end{figure}

\begin{figure*}[!thp]
  \centering
  \includegraphics[width=0.93\textwidth]{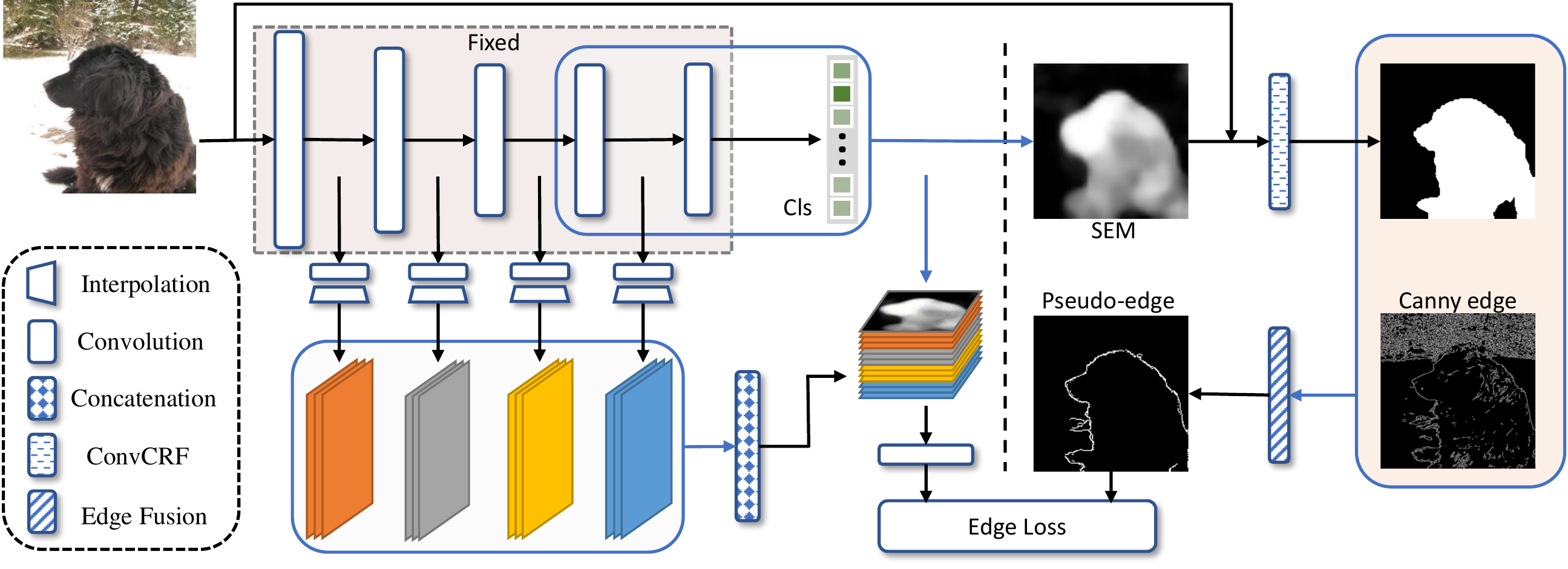}
  \caption{
  The network structure for training object boundaries.
  Pseudo-edges are generated based on the localization maps produced by SEM, and then employed as the supervision to train the boundaries with multiscale feature maps.
  }\label{fig-framework2}
  \vspace{-15pt}
\end{figure*}

\vspace{-1mm}
\subsection{Beyond Localization Map}\label{subsec-edge}

Previous works~\cite{zhou2015cnnlocalization,zhang2018adversarial,zhang2018self,singh2017hide} only aim at roughly locating the target objects rather than acquiring their fine details,~\ie, object boundaries. 
Here, we take a step further to explore if we can obtain much finer details of objects under the supervision of only image-level labels.
Different from fully supervised edge detection approaches~\cite{he2019bi,xie2015holistically} that require a large number of human-annotated edges for supervision, such detailed supervision information can hardly be obtained on such huge dataset (1.2M images).
Khoreva~\etal~\cite{khoreva2016weakly} proposes a weakly supervised method for detecting object boundaries.
However, this method requires bounding boxes of objects rather than only applying image-level labels.
It also uses intensive computational resources to infer the object boundaries using GrabCut~\cite{rother2004grabcut}, MCG~\cite{pont2016multiscale} and object detector~\cite{girshick15fastrcnn}, which prevents it from applying to a large-scale dataset.
To the best of our knowledge, our method is the first feasible attempt to conduct object boundary detection on such a large-scale dataset in a weakly supervised manner.

The framework of our SEM-based boundary detector is shown in Figure~\ref{fig-framework2}. 
Localization maps act as a vital bridge for learning boundaries.
We fix the network parameters of producing localization maps, and borrow the multi-scale features to predict boundaries.
Therefore, the quality of predicted boundaries can also reflect the fineness of discovering object details in localization maps.
Given an input image $I$, we generate $M_{sem}$ with the description in Section~\ref{subsec-scm}.
The maps can highlight the area of the target objects but fail to grasp the object boundaries.
To solve this issue, we first employ the map $M_{sem}$ as the unary potentials of ConvCRF~\cite{teichmann2018convolutional} to locate coarse object boundaries together with the input RGB image $I$.
Then, the located boundaries are fused with the Canny~\cite{canny1986computational} edges by finding the longest contour to generate pseudo-boundaries.
Finally, we use the pseudo-boundaries as supervision to train the boundary detector.
The boundary detector shares the backbone network with SEM and utilizes the multi-scale feature maps as the input.
Multi-scale feature maps are firstly fed into a convolution layer to adapt the channels, and then enlarged to predict boundaries.

Suppose $B$ denotes the ground-truth boundary mask, $B_{i,j}$ is $1$ if a pixel at $(i,j)$ is on boundaries, otherwise $B_{i,j}$ is $0$.
The typical loss function~\cite{xie2015holistically,liu2017richer} to optimize object boundaries is a cross-entropy function.
Due to the imbalance in the number between the boundary pixels and the background pixels, we apply the weights $\alpha=\frac{|B^{+}|}{|B^{+}|+|B^{-}|}$ and $\beta =\frac{|B^{-}|}{|B^{+}|+|B^{-}|}$ to balance the costs in the training loss, where $B^{+}$ and $B^{-}$ denote the positive and negative boundary pixels, respectively.
Therefore, the vanilla loss is as the second part in Eq.~\eqref{eq_vanilla_edge}.
\begin{equation}\label{eq_vanilla_edge}
\vspace{-2mm}
\small
\begin{aligned}
\vspace{-3mm}
\mathcal{L}^{e}(I) &=\frac{1}{|B^{+}|+|B^{-}|} \sum_{i,j}[\underbrace{\alpha (1-B_{i,j})P_{i,j}\log(1-P_{i,j})}_{HNS}\\
    &+\underbrace{\beta B_{i,j}\log~P_{i,j} 
+ \lambda\alpha (1-B_{i,j})\log(1-P_{i,j}))}_{vanilla} 
    ],
\end{aligned}
\vspace{-2mm}
\end{equation}
where $\lambda$ is a hyper-parameter.
$P_{i,j}$ is the predicted probability of being edge point at the pixel $(i, j)$.

The vanilla loss function can successfully recall edge pixels.
However, it fails in depressing false positive points.
The reason is that $\alpha$ is usually a very small value because the number of positive points only accounts for a very small portion of images.
If a false positive point occurs, the cost of this point will be negligible compared to the total cost. 
Enlarging the value of $\lambda$ could be a feasible way to address this problem, but it could harm the ability to recovering edge pixels.
To tackle this problem, we propose to add a Hard-Negative Suppression (HNS) item to eliminate the edges in background regions and only preserve the edges around target objects.
The proposed HNS item is shown in Eq.\eqref{eq_vanilla_edge}.
If the point $(i,j)$ mismatches with the pseudo-edge point, the score $P_{i,j}$ is expected to be 0.
For the hard-negative points with high scores $P_{i,j}$, they will cause high contributions to the total cost.
The higher hard-negative scores will induce higher costs, and therefore, their scores will be gradually suppressed with the training process.

\vspace{-2mm}
\section{Experiments}
\subsection{Experiment Setup}
\subsubsection{Datasets}
We mainly perform the experiments on two popular benchmarks,~\ie, ILSVRC~\cite{2009-imagenet} and CUB-200-2011~\cite{WahCUB_200_2011}, following the previous state-of-the-art methods~\cite{zhou2015cnnlocalization,zhang2018adversarial,zhang2018self}. ILSVRC is a widely acknowledged localization dataset including 1.2 million images of 1,000 categories for training and 50,000 images for validation. CUB-200-2011 includes 11,788 images (5,994 for training and 5,794 for testing) of 200 different species of birds.
In our experiments, the proposed method is learned on the training sets using only image-level labels as supervision.

\subsubsection{Evaluation Metrics}
\noindent \textbf{Direct Evaluation} We mainly utilize the proposed \textit{direct} metric to evaluate the localization performance by performing a pixel-wise comparison between the produced localization map and its corresponding ground-truth mask.
For ILSVRC, we annotate the ground-truth masks for the images on the validation set.
Particularly, we first manually exclude 5,729 images that have ambiguous object pixels and  annotate the left 44,271 images with an interactive object segmentation approach~\cite{zhang2020cvpr}.
Then, the annotated masks are split for validation (23,151 images) and test (21,120 images), respectively. 
Please see the supplementary material for more details of the annotated masks.
For CUB-200-2011, the ground-truth masks have already been provided in the original dataset. Given the produced localization maps, each pixel is with probability value bounded in $[0,1]$, and the desired object-related pixels are expected to have high activation values. For the convenience of visualization and evaluation, we map the values to the range of $[0,255]$. During evaluation, we binarize the localization maps using a threshold $T\in [0,255]$, and calculate IoU scores of the binarized masks against ground-truth masks \wrt different thresholds. 
We mainly compare different algorithms under the IoU-Threshold curve and its two key properties,~\ie, Peak-IoU and Peak-T.
In addition, we also report the Precision-Recall curve and the Average-Precision (AP) scores.

\noindent \textbf{Indirect Evaluation} Following previous approaches~\cite{zhou2015cnnlocalization, zhang2018adversarial,zhang2018self}, we also conduct the evaluation using the \textit{indirect} measurement to illustrate the superiority of the proposed SEM method. To be specific, this metric calculates the percentage of the images that can satisfy the following two conditions simultaneously.
1) The predicted classification labels match the ground-truth categories;
2) The predicted bounding boxes have over 50\% IoU with at least one of the ground-truth boxes. 
This criterion involves two aspects,~\ie, classification accuracy and localization accuracy.
We also evaluate the localization maps \wrt the ground-truth labels, which is dubbed as Gt-known accuracy.

\subsubsection{Implementation Details}
To understand the true localization ability of localization maps produced by the current methods, we first implement several recent benchmark methods,~\ie, CAM~\cite{zhou2015cnnlocalization}, ACoL~\cite{zhang2018adversarial}, SPG~\cite{zhang2018self}, ADL~\cite{Choe_2019_CVPR} and CutMix~\cite{yun2019cutmix}.
We apply the same backbone network,~\ie, InceptionV3~\cite{szegedy2016rethinking}, and the same training strategy for a fair comparison.
We train the networks following the way in the original papers~\cite{zhou2015cnnlocalization,zhang2018adversarial,zhang2018self,yun2019cutmix,Choe_2019_CVPR,singh2017hide}.
During testing, We use their own localization maps as the fist-stage guidance and apply the proposed SEM method to enhance localization maps.
If there is no specific mentioning, SEM is applied on the CAM~\cite{zhou2015cnnlocalization} method using InceptionV3 by default in this paper.
To fully understand the proposed SEM in terms of different backbone networks, we further implement the proposed SEM based on various popular backbone networks, including VGG16~\cite{simonyan2014very}, InceptionV3~\cite{szegedy2016rethinking} and ResNets~\cite{he2016deep}.
We follow the baseline methods,~\eg, CAM~\cite{zhou2015cnnlocalization}, ACoL~\cite{zhang2018adversarial} and SPG~\cite{zhang2018self}, to adapt the backbone classification networks for producing decent localization maps.
In particular, we remove the fully connected layer on the top and change the last two $stride=2$ to $stride=1$ for enlarging the size of feature maps. 
We add a block of three convolution layers to adjust channel numbers, which follows ACoL~\cite{zhang2018adversarial} to obtain final localization maps.
For the prediction of SEM-Edge, we simply borrow the feature maps from the backbone networks.
Explicitly, the feature maps from the first to fourth blocks are adjusted to the channel size of $128$ by a \textit{$Conv3\times 3$-BN-ReLU} layer, and then upsampled to the size of $(160,160)$.
We concatenate the multi-scale feature maps and the localization maps produced by SEM, and then feed these features into three convolution layers to predict the final edge maps.
We follow the training practices of the previous works~\cite{zhang2018adversarial, zhang2018self}.
The networks are initialized by the weights trained on ILSVRC.
For the training process on ILSVRC, the networks are fine-tuned for five epochs with the initial learning rate of $0.001$.
The learning rate is decreased by a factor of 10 after the second epoch.
The learning rate of the added top three convolutional layers is $10$ times of the above configuration.


\begin{figure*}[t]
  \centering
    \subfloat[IoU-Threshold on val\label{fig:curves-baselines-a}]{\includegraphics[width=0.246\textwidth]{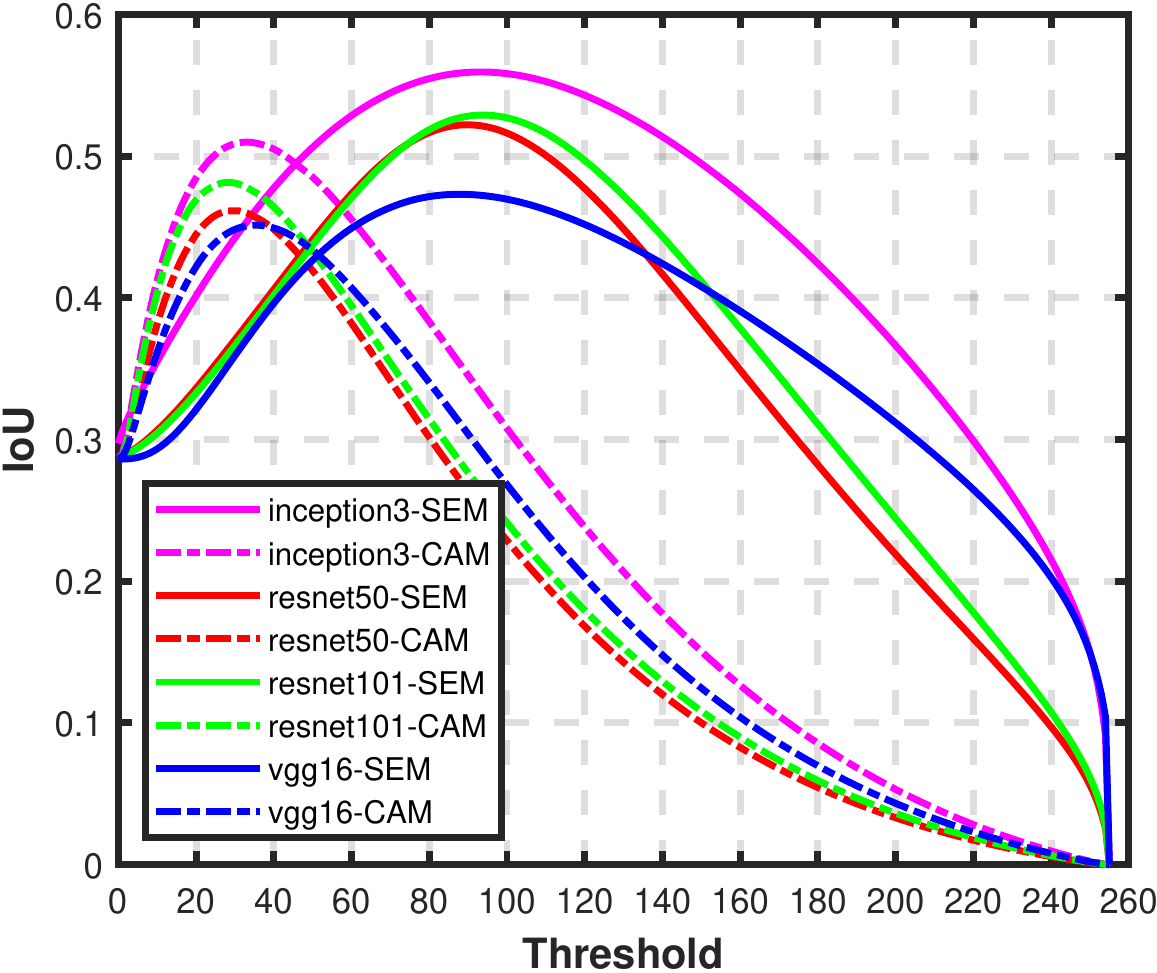}}
    \hspace{1pt}
    \subfloat[IoU-Threshold on test\label{fig:curves-baselines-b}]{\includegraphics[width=0.245\textwidth]{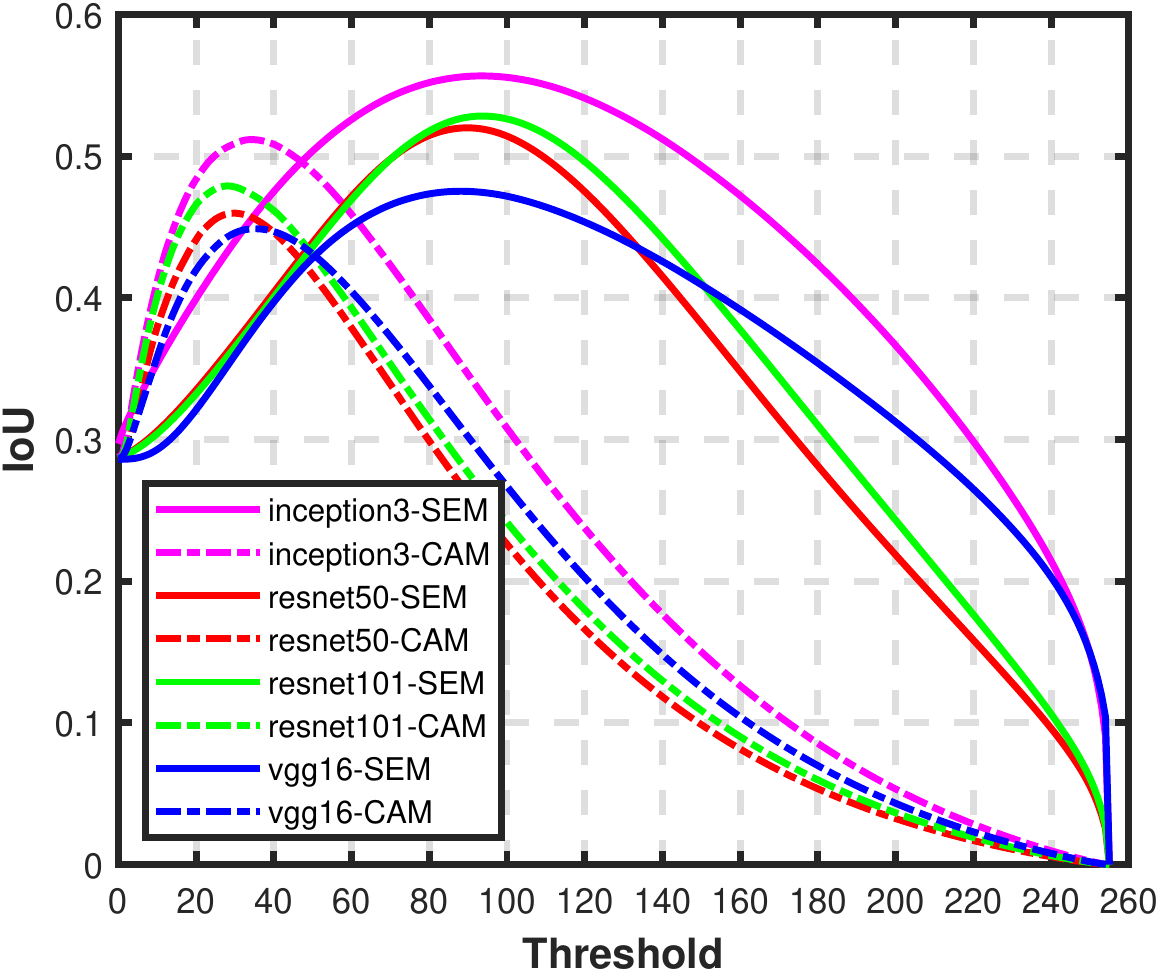}}
    \hspace{1pt}
    \subfloat[Precision-Recall on val\label{fig:curves-baselines-c}]{\includegraphics[width=0.24\textwidth]{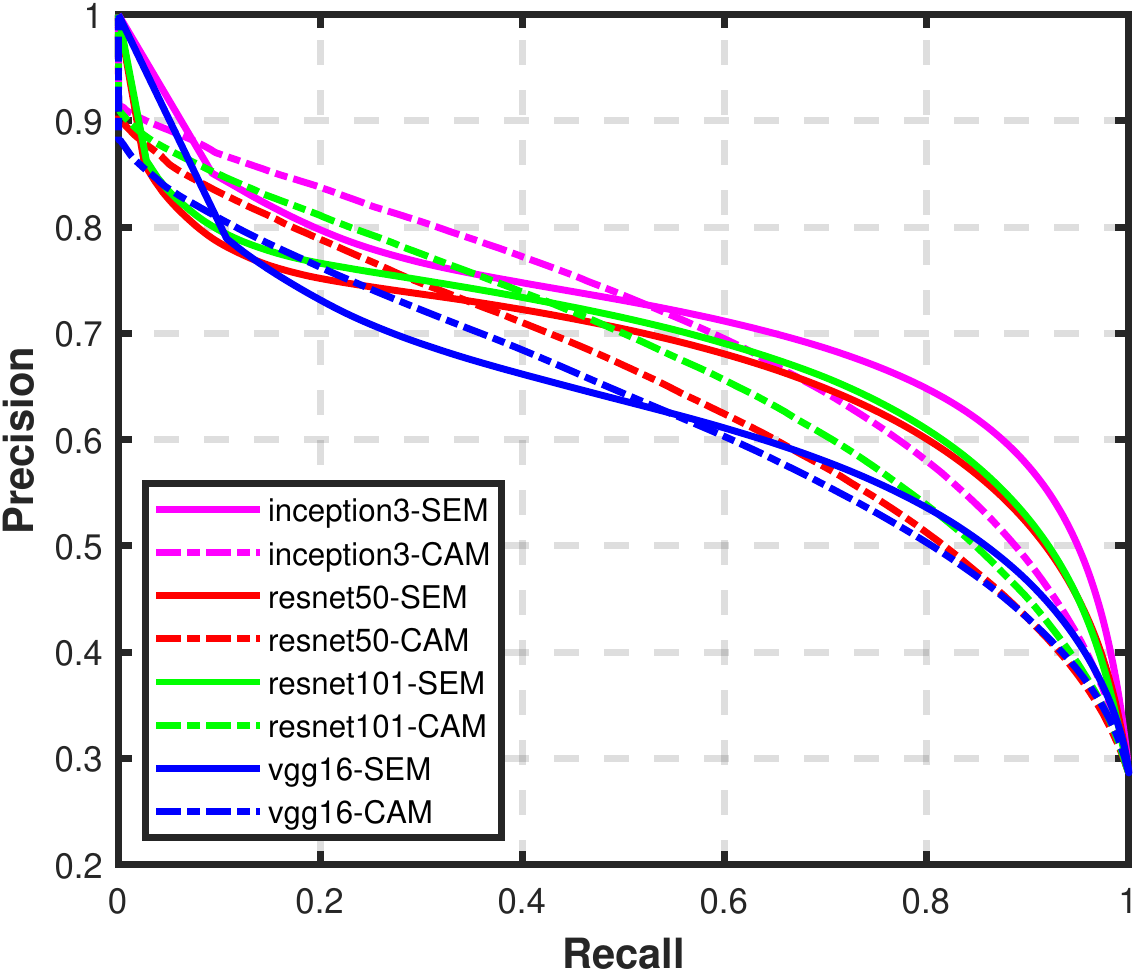}}
    \hspace{1pt}
    \subfloat[Precision-Recall on test\label{fig:curves-baselines-d}]{\includegraphics[width=0.24\textwidth]{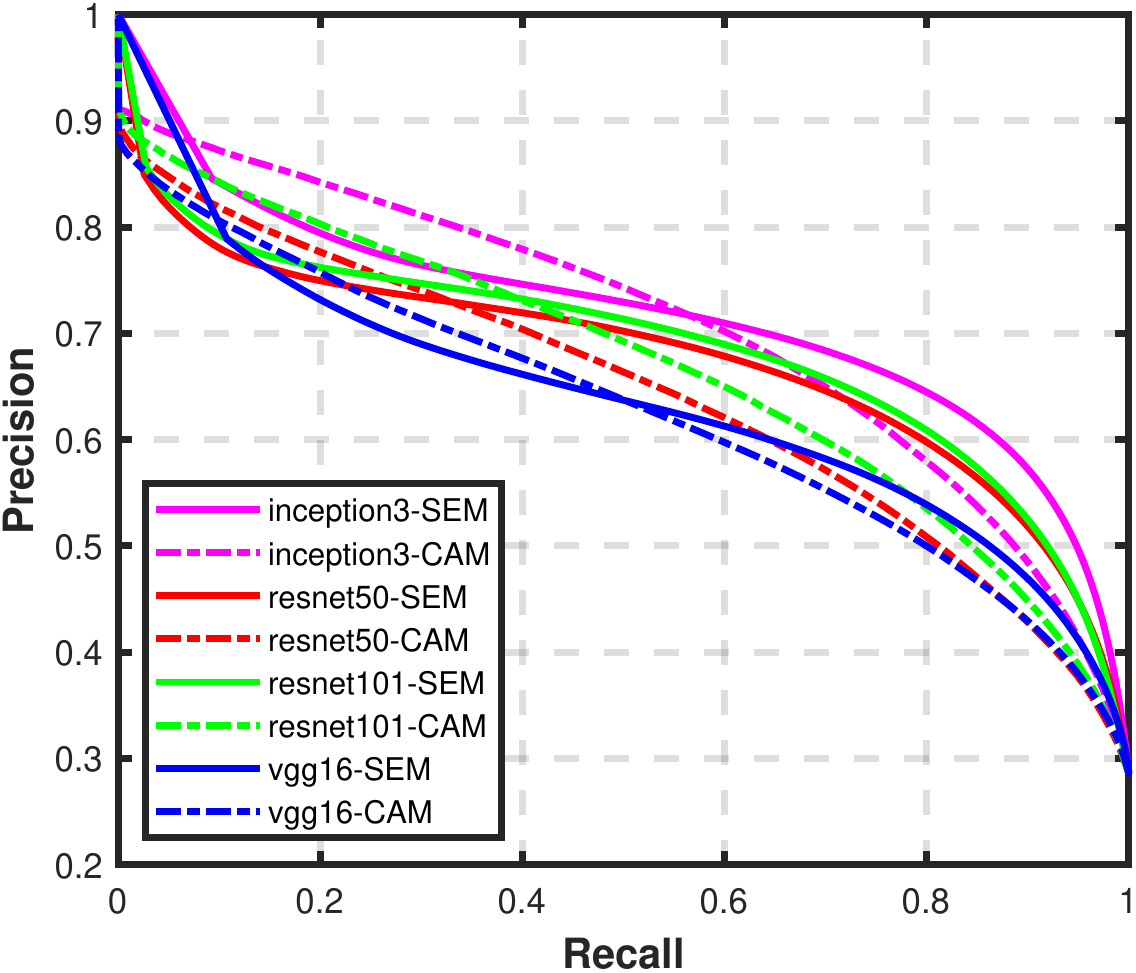}}
    
  \vspace{-5pt}
  \caption{
  The IoU-Threshold and Precision-Recall curves with different backbone networks on the annotated splits of ILSVRC.
  }\label{fig:curves-baselines}
  \vspace{-5pt}
\end{figure*}

\begin{figure*}[t]
  \centering
    \subfloat[IoU-Threshold on val\label{fig:curves-backbones-a}]{\includegraphics[width=0.246\textwidth]{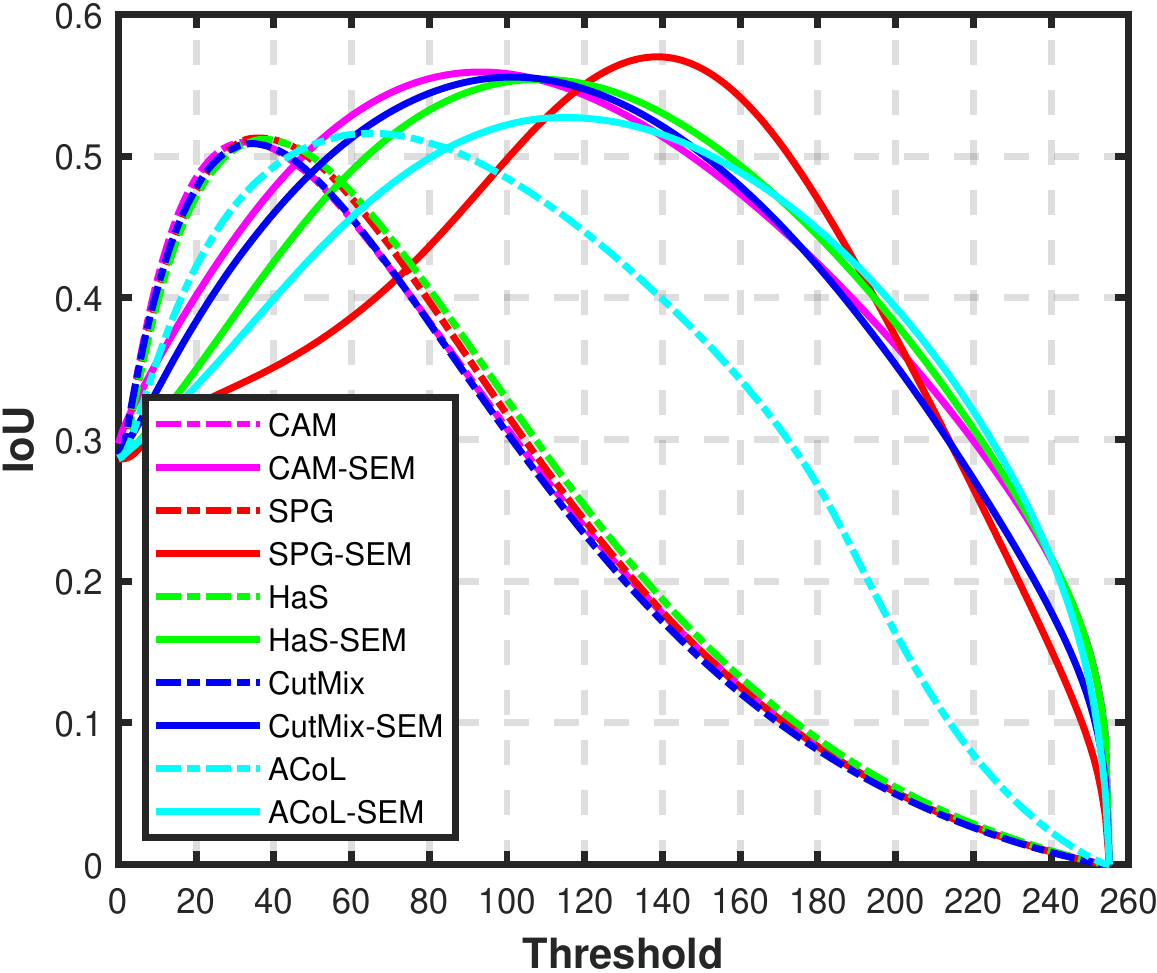}}
    \hspace{1pt}
    \subfloat[IoU-Threshold on test\label{fig:curves-backbones-b}]{\includegraphics[width=0.245\textwidth]{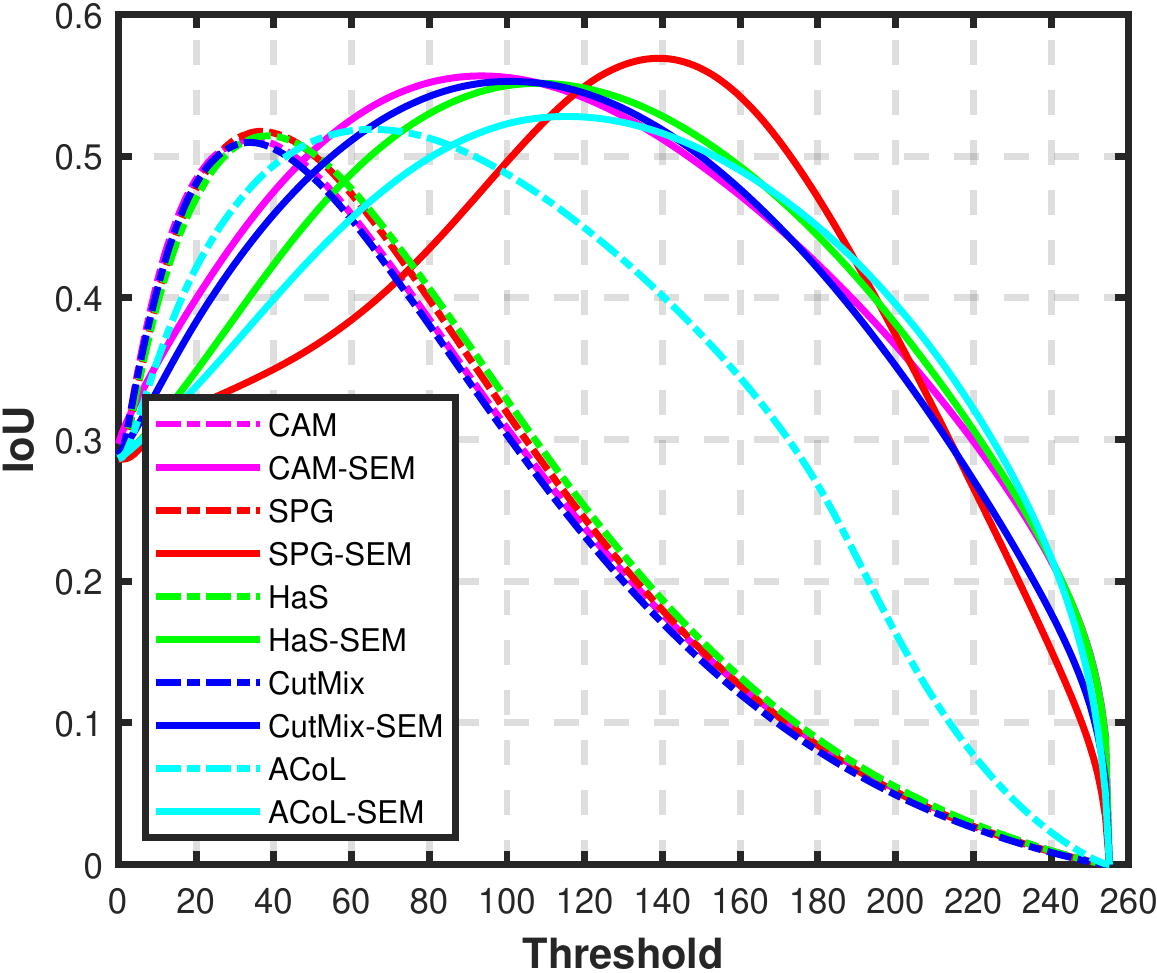}}
    \hspace{1pt}
    \subfloat[Precision-Recall on val\label{fig:curves-backbones-c}]{\includegraphics[width=0.24\textwidth]{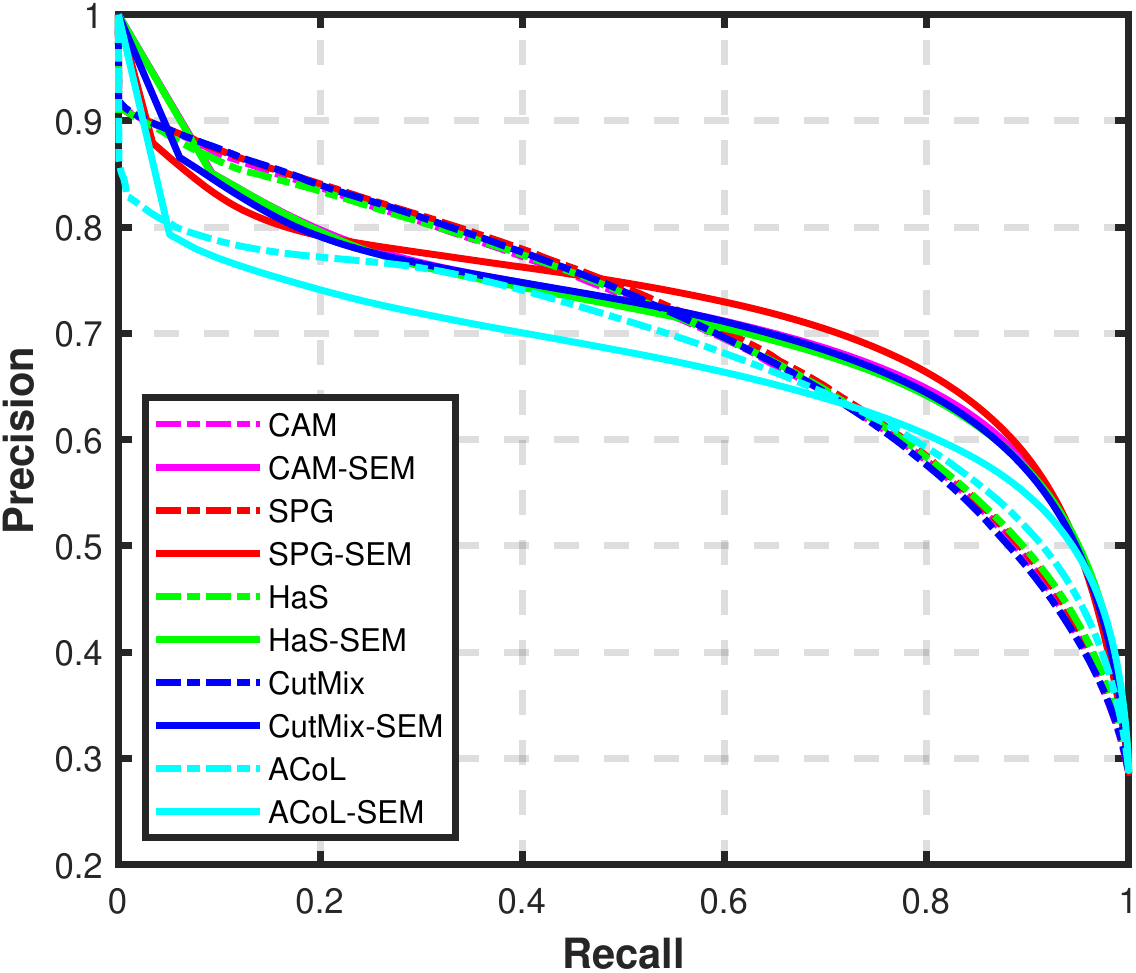}}
    \hspace{1pt}
    \subfloat[Precision-Recall on test\label{fig:curves-backbones-d}]{\includegraphics[width=0.24\textwidth]{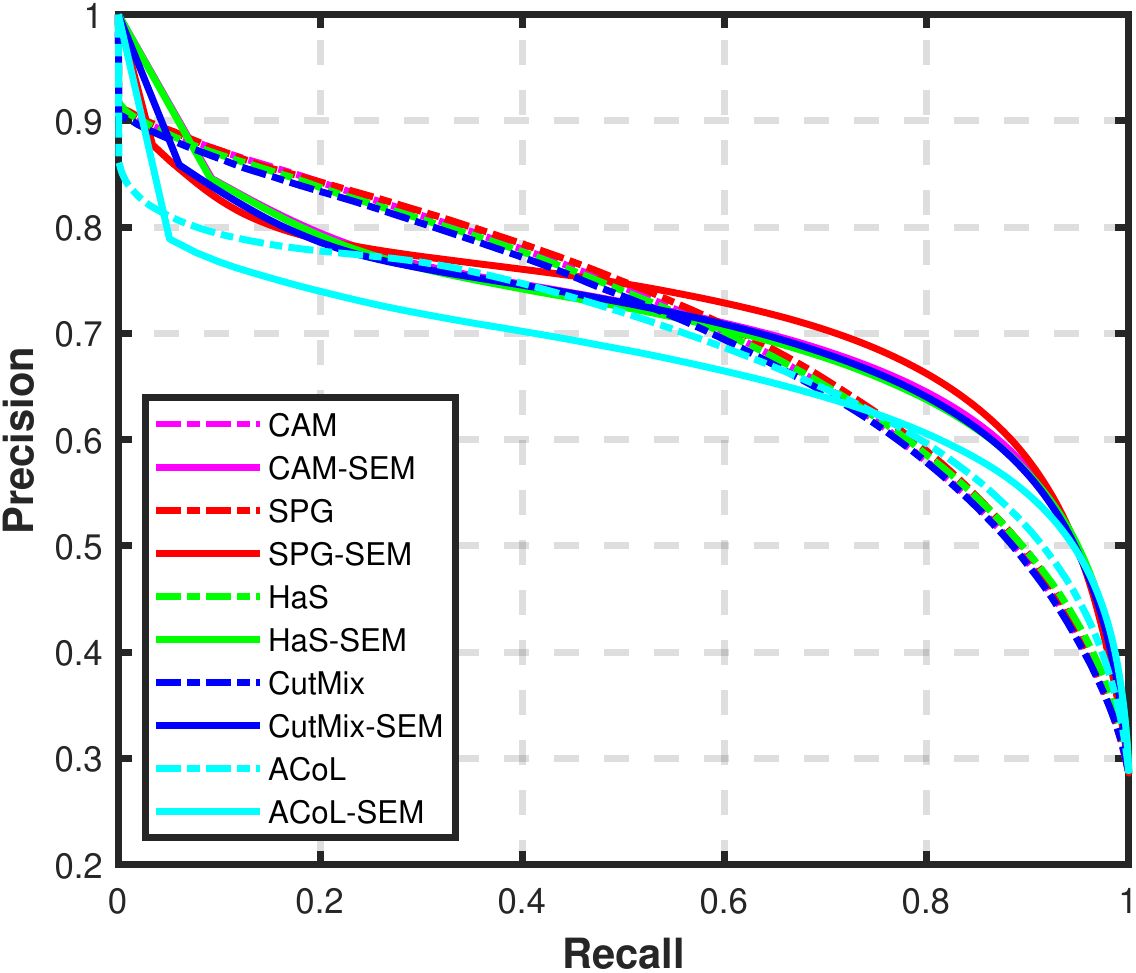}}
    
  \vspace{-5pt}
  \caption{
  The IoU-Threshold and Precision-Recall curves of different baseline methods on the annotated splits of ILSVRC.
  }\label{fig:curves-backbones}
  \vspace{-5pt}
\end{figure*}

\begin{figure}[t]
  \vspace{-10pt}
  \centering
    \subfloat[IoU-Threshold\label{fig:cub-backbones-a}]{\includegraphics[width=0.243\textwidth]{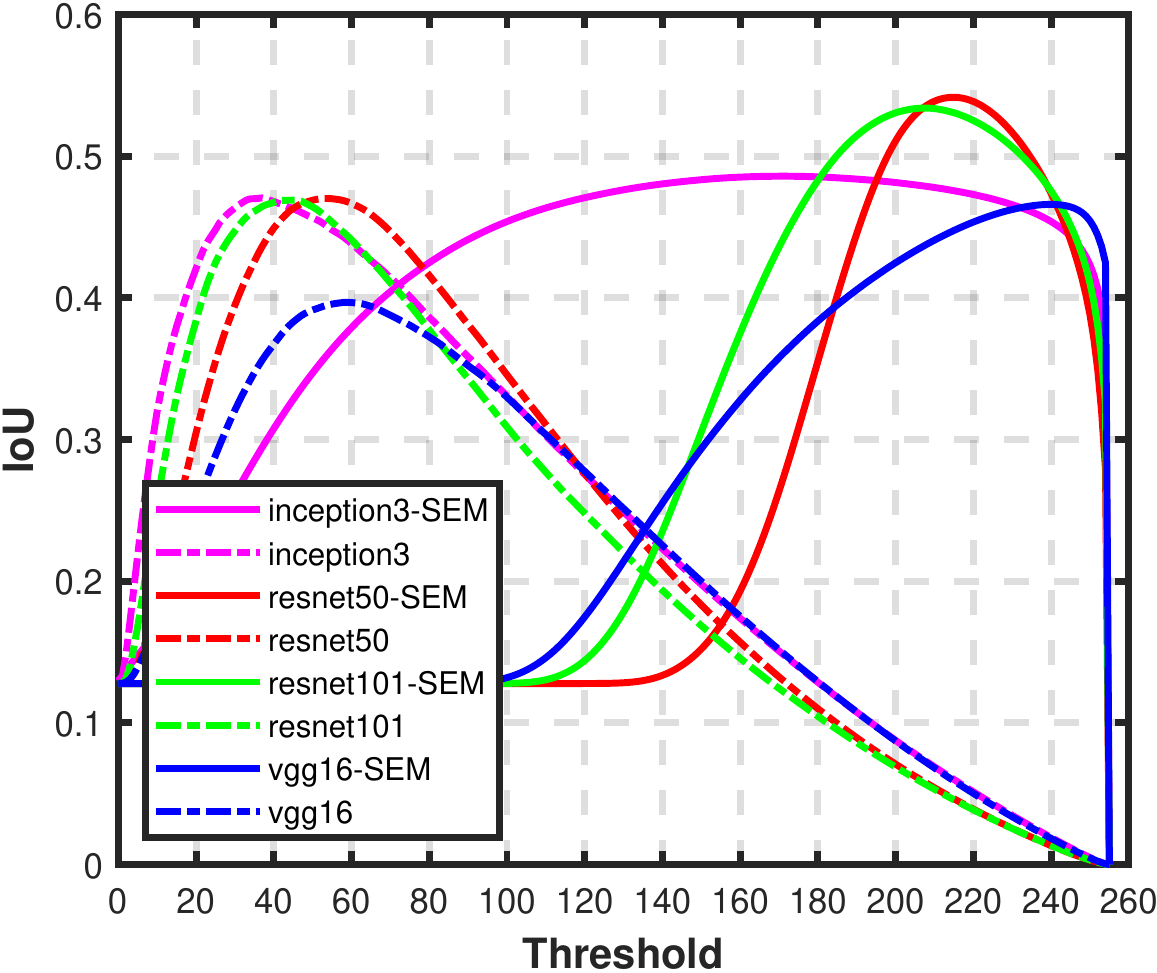}}
    \hspace{-1pt}
    \subfloat[Precision-Recall\label{fig:cub-backbones-b}]{\includegraphics[width=0.24\textwidth]{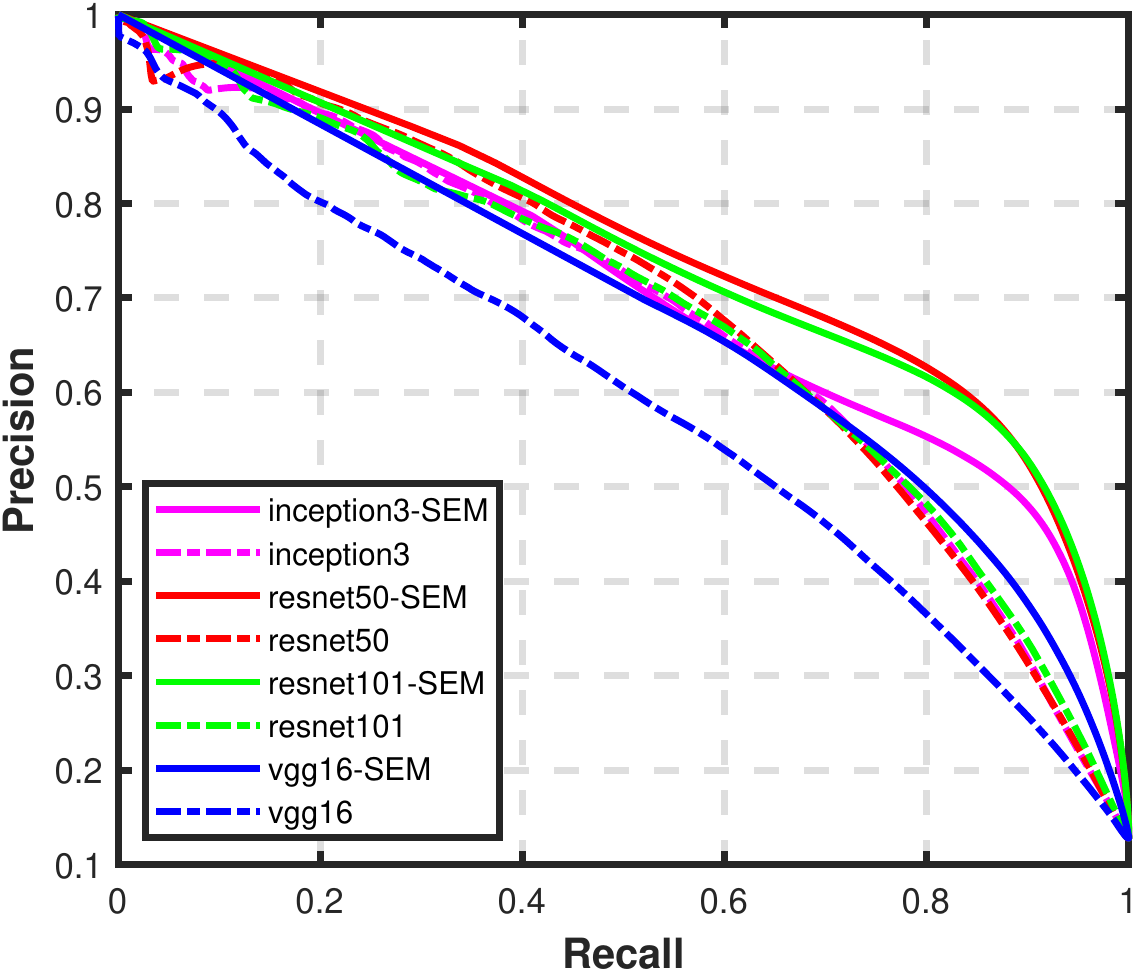}}
    
  \caption{
  The IoU and Precision-Recall curves of different backbone networks on the test set of CUB.
  }\label{fig:cub-backbones}
  \vspace{-15pt}
\end{figure}

\subsection{Comparison with the state-of-the-arts}
First, we evaluate and compare the recent baseline methods in the proposed \textit{direct} metric.
We adapt our training-free SEM method into these baselines to illustrate the effectiveness.
We also conduct various backbone networks to verify the robustness of SEM.
Second, we show that the proposed SEM can also boost the accuracy of the inferred bounding boxes in the traditional \textit{indirect} measurement.
Our improved localization maps achieve the best accuracy and significantly surpass the existing methods. 

\begin{figure}[htbp]
  \centering
  \includegraphics[width=0.49\textwidth]{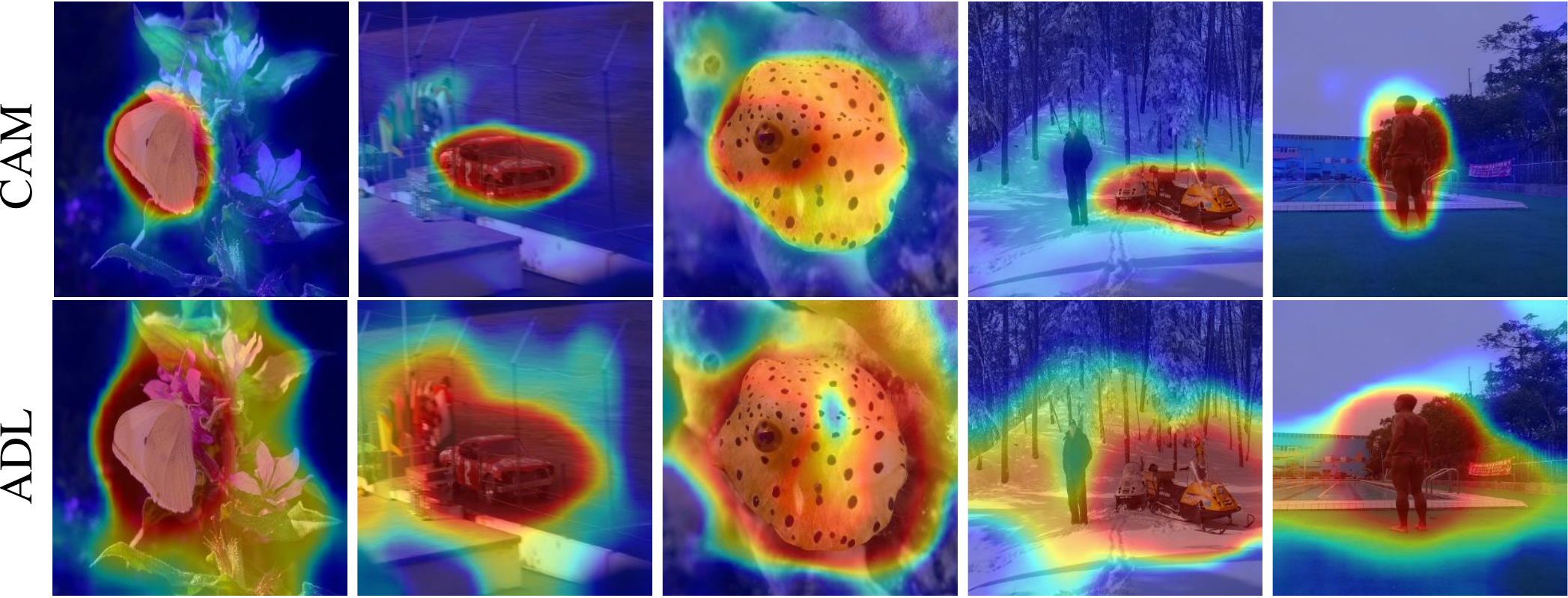}
  \caption{
  Localization maps extracted by SEM.
  When using ADL, background noises around the target objects are incorporated caused by the erasing operation on feature maps.
  }\label{fig-failure}
  \vspace{-20pt}
\end{figure}

\noindent \textbf{Peak-IoU}
Table~\ref{tab:comp-baselines} shows the recent weakly localization approaches under the proposed \textit{direct} evaluation metric.
We use ``w/" and ``w/o" to indicate whether the proposed SEM is applied to enhance the quality of localization maps.
The proposed SEM successfully surpasses the counterparts in the proposed direct metric of Peak-IoU and Peak-T.
We can observe that Peak-IoU values of CAM are 50.58\% and 50.13\% on our annotated validation and test sets, respectively.
The other methods,~\ie, HaS, ACoL, CutMix, SPG and ADL, which are proposed to improve the quality of localization maps based on CAM, have increased the Peak-IoU value only by 1.01\% and 1.77\% (ACoL) at most.
Not all of these methods have really succeeded in improving localization maps.
By contrast, the proposed SEM module achieves consistent improvement over every approach.
The Peak-IoU values increase to 57.02\% and 56.92\% (SPG w/ SEM) on the validation and testing sets, respectively.
Our SEM method successfully achieves a new state-of-the-art accuracy, surpassing the best values w/o SEM by 5.43\% and 5.02\% on the two splits. 
Compared to the method w/o SEM, SEM achieves the most significant improvement on SPG~\cite{zhang2018self}, increasing Peak-IoU by 5.74\% and 5.16\% on the two splits, respectively. 
ACoL~\cite{zhang2018adversarial} and ADL~\cite{Choe_2019_CVPR}, which apply erasing operations on high-level feature maps to force the networks to find more object regions, obtain the least increase.
The cause for the limited improvements of these two methods is that the erasing operation on feature maps may include background noises.
We show the included noises in Figure~\ref{fig-failure}.


We also validate the robustness of the proposed SEM method on various backbone networks.
In Table~\ref{tab:comp-backbones}, we choose four different popular backbone networks, and compare our SEM model with the CAM method.
SEM obtains the smallest improvement on VGG16 of increasing the Peak-IoU values by 2.19\% and 2.64\%, while achieves the largest improvement on ResNet50 by 6.08\% and 6.03\% on the validation and test splits, respectively.
Also, we validate the SEM method on the CUB-200 dataset upon different backbone networks.
All of the Peak-IoU results surpass the CAM counterparts.
ResNet50 also achieves the largest gain of rising by 7.14\% from 47.02\% to 54.16\%.

We draw the IoU-Threshold curves to give an even clear comparison of various baseline methods in Figure~\ref{fig:curves-baselines-a} and ~\ref{fig:curves-baselines-b}.
It is easy to read from the curves that 1) the baseline methods using CAM to extract localization maps have lower Peak-IoU values, while the Peak-IoU values obviously increase after applying the proposed SEM module; 
2) the Peak-T values drift towards bigger values after applying SEM, reflecting that the localization maps have better visual effects. 
Figure~\ref{fig:curves-backbones-a} and ~\ref{fig:curves-backbones-b} illustrate the IoU-Threshold curves regarding various backbone networks.
We can also observe that the SEM method can not only improve Peak-IoU values but promote the Peak-T to produce better localization maps for visualization.

\newcolumntype{g}{>{\columncolor[gray]{0.9}}c}
\begin{table*}[t]\setlength{\tabcolsep}{13pt}
  \caption{
  Comparison of different methods for extracting localization maps under the amended \emph{direct} evaluation metric on ILSVRC. w/o and w/ indicate whether the proposed SEM is applied for enhancement.
  Better results are in \textbf{bold}.
  }\label{tab:comp-baselines}
  \vspace{-5pt}
  \centering
  \small
  \begin{tabular}{lcgggggg}
     \toprule[0.2em]
    \multirow{2}{*}{Method} & 
    \multirow{2}{*}{SEM} & 
    \multicolumn{2}{c}{AP} & \multicolumn{2}{c}{Peak-IoU} &
    \multicolumn{2}{c}{Peak-T} \\
    \cline{3-8}
    & & \cellcolor{white}val & \cellcolor{white}test & \cellcolor{white}val & \cellcolor{white}test & \cellcolor{white}val & \cellcolor{white}test\\
     \toprule[0.2em]
    \rowcolor{white}
     \multirow{2}{*}{CAM~\cite{zhou2015cnnlocalization}}
     & w/o & 70.68 & 71.06 & 50.58 & 50.13 & 33 & 38 \\
     & \cellcolor[gray]{0.9}w/ & \textbf{72.78}~(+2.10) & \textbf{72.54}~(+1.48) & \textbf{56.05}~(+5.05) & \textbf{55.67}~(+5.54)  & \textbf{93} & \textbf{93} \\
     \bottomrule[0.1em]
    \rowcolor{white}
     \multirow{2}{*}{HaS~\cite{singh2017hide}}
      & w/o & 70.79 & 71.07 & 51.20 & 51.44 & 37 & 38 \\
      & \cellcolor[gray]{0.9}w/ & \textbf{72.46}~(+1.67) & \textbf{72.24}~(+1.17)  & \textbf{55.43}~(+4.23) & \textbf{55.15}~(+3.71) & \textbf{109} & \textbf{108}\\
     \bottomrule[0.1em]
    \rowcolor{white}
     \multirow{2}{*}{ACoL~\cite{zhang2018adversarial}}
      & w/o  & \textbf{68.15} & \textbf{68.64} & 51.59 &51.90 & 62 &65 \\
      & \cellcolor[gray]{0.9}w/  & 67.89~(-0.26) & 67.92~(-0.72) & \textbf{52.74}~(+0.84) & \textbf{52.80}~(+0.90)& \textbf{115} & \textbf{115} \\
     \bottomrule[0.1em]
    \rowcolor{white}
     \multirow{2}{*}{CutMix~\cite{yun2019cutmix}}
     & w/o  & 70.78 & 70.45 & 50.89 &50.97 & 34 &34 \\
     & \cellcolor[gray]{0.9}w/  & \textbf{72.12}~(+1.34) & \textbf{71.78}~(+1.33) & \textbf{55.59}~(+4.70) & \textbf{55.28}~(+4.31) & \textbf{101} & \textbf{101} \\
     \bottomrule[0.1em]
    \rowcolor{white}
     \multirow{2}{*}{SPG~\cite{zhang2018self}}
     & w/o  & 71.18 & 71.53 & 51.28& 51.76 & 36 & 37  \\
     & \cellcolor[gray]{0.9}w/  & \textbf{72.77}~(+1.59) & \textbf{72.59}~(+1.06) & \textbf{57.02}~(+5.74) & \textbf{56.92}~(+5.16) & \textbf{139} & \textbf{139}  \\
     \bottomrule[0.1em]
    \rowcolor{white}
     \multirow{2}{*}{ADL~\cite{Choe_2019_CVPR}}
     & w/o  & \textbf{69.89} & \textbf{70.19} & 50.01 & 49.96 & 42 & 42  \\
     & \cellcolor[gray]{0.9}w/  & 67.32~(-2.57) & 67.26~(-2.93) & \textbf{51.10}~(+1.09) & \textbf{50.98}~(+1.02) & \textbf{144} & \textbf{144}  \\
     \bottomrule[0.1em]
  \end{tabular}
  \vspace{-10pt}
\end{table*}
\begin{table*}[t]\setlength{\tabcolsep}{13pt}
  \caption{
  Comparison of different backbone networks for extracting localization maps with CAM~\cite{zhou2015cnnlocalization} under the amended \emph{direct} evaluation metric on ILSVRC. w/o and w/ indicate whether the proposed SEM is applied for enhancement.
  Better results are in \textbf{bold}.
  }\label{tab:comp-backbones}
  \vspace{-5pt}
  \centering
  \small
  \begin{tabular}{lcgggggg}
     \toprule[0.2em]
    \multirow{2}{*}{Backbone} & 
    \multirow{2}{*}{SEM} & 
    \multicolumn{2}{c}{AP} & \multicolumn{2}{c}{Peak-IoU} &
    \multicolumn{2}{c}{Peak-T} \\
    \cline{3-8}
     & & \cellcolor{white}val & \cellcolor{white}test & \cellcolor{white}val & \cellcolor{white}test & \cellcolor{white}val & \cellcolor{white}test\\
     \toprule[0.2em]
     \rowcolor{white}
     \multirow{2}{*}{VGG16~\cite{zhou2015cnnlocalization}}
     & w/o & 63.30 & 62.82 & 45.13 & 44.89 & 35 & 35 \\
     & \cellcolor[gray]{0.9}w/ & \textbf{65.03}~(+1.73) & \textbf{65.14}~(+2.32) & \textbf{47.32}~(+2.19) & \textbf{47.53}~(+2.64)  & \textbf{89} & \textbf{88} \\
     \bottomrule[0.1em]
     \rowcolor{white}
     \multirow{2}{*}{ResNet50~\cite{singh2017hide}}
      & w/o & 65.18 & 64.51 & 46.14 & 45.98 & 31 & 30 \\
      & \cellcolor[gray]{0.9}w/ & \textbf{68.12}~(+2.94) & \textbf{67.83}~(+1.69)  & \textbf{52.22}~(+6.08) & \textbf{52.01}~(+6.03) & \textbf{90} & \textbf{90}\\
     \bottomrule[0.1em]
     \rowcolor{white}
     \multirow{2}{*}{ResNet101~\cite{zhang2018adversarial}}
      & w/o  & 67.56 & 66.99 & 48.16 & 47.91 & 28 & 28 \\
      & \cellcolor[gray]{0.9}w/  & \textbf{69.08}~(+1.52) & \textbf{68.87}~(+1.88) & \textbf{52.91}~(+4.75) & \textbf{52.84}~(+4.93)& \textbf{94} & \textbf{94} \\
     \bottomrule[0.1em]
     \rowcolor{white}
     \multirow{2}{*}{Inception3~\cite{yun2019cutmix}}
     & w/o  & 70.68 & 71.06 & 51.00 & 51.17 & 33 & 38 \\
     & \cellcolor[gray]{0.9}w/  & \textbf{72.78}~(+2.10) & \textbf{72.54}~(+1.48) & \textbf{56.05}~(+5.05) & \textbf{55.67}~(+4.50) & \textbf{93} & \textbf{93} \\
     \bottomrule[0.1em]
  \end{tabular}
  \vspace{-10pt}
\end{table*}
\newcolumntype{g}{>{\columncolor[gray]{0.9}}c}
\begin{table}[t]\setlength{\tabcolsep}{4pt}
  \caption{
  Comparison with different backbone networks on CUB.
  w/o of SEM refers to using CAM~\cite{zhou2015cnnlocalization} for extracting localization maps, otherwise using the proposed SEM.
  }
  \label{tab:comp-cub-nets-sal}
  \vspace{-10pt}
  \centering
  \small
  \begin{tabular}{lgggg}
     \toprule[0.2em]
     \rowcolor{white}
     Backbone & SEM & AP & Peak-IoU & Peak-T\\
     \toprule[0.2em]
     \rowcolor{white}
     \multirow{2}{*}{VGG16~\cite{simonyan2014very} }
     & w/o  &  59.17 & 39.69 &  49 \\
     & w/ & \textbf{76.29}(+17.12) & \textbf{46.60}(+6.91) & \textbf{240}\\
     \hline
     \rowcolor{white}
     \multirow{2}{*}{ResNet50~\cite{he2016deep} }
     & w/o  &  68.90 & 47.02 & 53 \\
     & w/ & \textbf{77.46}(+8.56) & \textbf{54.16}(+7.14) & \textbf{215} \\
     \hline
     \rowcolor{white}
     \multirow{2}{*}{ResNet101~\cite{he2016deep}}
     & w/o  & 68.42 & 46.91 & 46\\
     & w/ & \textbf{77.45}(+9.03) & \textbf{53.40}(+6.49) & \textbf{208} \\
     \hline
     \rowcolor{white}
     \multirow{2}{*}{InceptionV3~\cite{szegedy2016rethinking}}
     & w/o &  68.11 & 47.06 & 37\\
     & w/ & \textbf{75.63}(+7.52) & \textbf{48.59}(+1.53) & \textbf{171} \\
     \bottomrule[0.1em]
  \end{tabular}
  \vspace{-15pt}
\end{table}
\begin{table}[t]\setlength{\tabcolsep}{5pt}
  \centering
  \caption{
  Localization accuracy of the inferred bounding boxes on the ILSVRC validation set. {$\dagger$}: The results of our re-implemented model.
  }\label{tab-comp-ilsvrc}
  \vspace{-10pt}
  \footnotesize
  \begin{tabular}{llccccc}
     \toprule[0.2em]
     \multirow{2}{*}{Backbone} & \multirow{2}{*}{Method} & \multicolumn{2}{c}{Loc. Acc.} & Cls. Acc.\\
      & & Top-1  & Gt-known & Top-1 \\
     \toprule[0.2em]
     \multirow{6}{*}{VGG16~\cite{simonyan2014very}}
     & {$\dagger$}CAM~\cite{zhou2015cnnlocalization}  & 45.15 & 60.06 &  71.2 \\
     &   & 42.80~\cite{zhou2015cnnlocalization} & - &  68.8~\cite{zhou2015cnnlocalization} \\
     & ADL~\cite{Choe_2019_CVPR} & 44.92 & - & 67.8  \\
     & ACoL~\cite{zhang2018adversarial}  & 45.83 & 62.96 & 67.5 \\
     & DFM~\cite{zhou2019dual} & 47.41 & - & 68.6 \\
     & \cellcolor[gray]{0.9}SEM~\textsubscript{CAM} & \cellcolor[gray]{0.9}\textbf{47.53} & \cellcolor[gray]{0.9}\textbf{63.47} & \cellcolor[gray]{0.9}71.2 \\
     \hline
     \multirow{3}{*}{ResNet50~\cite{he2016deep}}
     & {$\dagger$}CAM~\cite{zhou2015cnnlocalization}  &  51.13 & 62.71 & 75.1 \\
     & DFM~\cite{zhou2019dual} & 49.71 & - & 77.8 \\
     & \cellcolor[gray]{0.9}SEM~\textsubscript{CAM} & \cellcolor[gray]{0.9}\textbf{53.84} & \cellcolor[gray]{0.9}\textbf{67.00} & \cellcolor[gray]{0.9}75.1 \\
     \hline
     \multirow{3}{*}{ResNet101~\cite{he2016deep}}
     & {$\dagger$}CAM~\cite{zhou2015cnnlocalization}  & 52.94 & 64.13 & 79.1    \\
     & DFM~\cite{zhou2019dual} & 50.67 & - & 77.2 \\
     & \cellcolor[gray]{0.9}SEM~\textsubscript{CAM} &  \cellcolor[gray]{0.9}\textbf{54.88} & \cellcolor[gray]{0.9}\textbf{67.15} & \cellcolor[gray]{0.9}79.1 \\
     \hline
     \multirow{9}{*}{InceptionV3~\cite{szegedy2016rethinking}}
     & {$\dagger$}CAM~\cite{zhou2015cnnlocalization} &  50.20 & 65.05 & 73.3 \\
     & SPG~\cite{zhang2018self}  & 48.60  & 64.69 & 69.7 \\
     & ADL~\cite{Choe_2019_CVPR}  & 48.71 & - & 72.8 \\
     & {$\dagger$}HaS & 50.08 & 65.66 & 71.7 \\
     & {$\dagger$}CutMix & 47.99 & 64.46 & 69.8 \\
     & \cellcolor[gray]{0.9}SEM~\textsubscript{HaS} & \cellcolor[gray]{0.9}51.59 & \cellcolor[gray]{0.9}68.38 & \cellcolor[gray]{0.9}71.7 \\
     & \cellcolor[gray]{0.9}SEM~\textsubscript{CutMix} & \cellcolor[gray]{0.9}50.79 & \cellcolor[gray]{0.9}68.98 & \cellcolor[gray]{0.9}69.8\\
     & \cellcolor[gray]{0.9}SEM~\textsubscript{SPG} & \cellcolor[gray]{0.9}50.79 & \cellcolor[gray]{0.9}\textbf{69.26} & \cellcolor[gray]{0.9}69.7 \\
     & \cellcolor[gray]{0.9}SEM~\textsubscript{CAM} & \cellcolor[gray]{0.9}\textbf{53.04} & \cellcolor[gray]{0.9}69.04 & \cellcolor[gray]{0.9}73.3\\
     \toprule[0.1em]
  \end{tabular}
  \vspace{-15pt}
\end{table}

\begin{table}[t]\setlength{\tabcolsep}{16pt}
  \centering
  \caption{Top-1 localization accuracy of the inferred bounding boxes on CUB based on InceptionV3~\cite{szegedy2016rethinking}.}\label{tab:comp-cub}
  \vspace{-10pt}
  \begin{tabular}{cccc}
    \toprule[0.2em]
    CAM~\cite{zhou2015cnnlocalization} & SPG~\cite{zhang2018self} & ADL~\cite{Choe_2019_CVPR}  & \cellcolor[gray]{0.9}SEM\\
    \toprule[0.2em]
    43.67 & 46.64 & 53.04 & \cellcolor[gray]{0.9}\textbf{61.57} \\
    \toprule[0.1em]
  \end{tabular}
  \vspace{-15pt}
\end{table}

\noindent \textbf{Precision-Recall}
Beyond the proposed IoU-Threshold metric, we also evaluate the predicted localization maps against the ground-truth masks using Precision-Recall curves and Average Precision (AP) values.
In Table~\ref{tab:comp-baselines}, we first see the typical CAM method achieves AP of 70.68\% and 71.06\% on the annotated validation and test sets, respectively.
It is noticeable that only HaS and SPG obtain consistent improvement on the two splits, while other methods,~\ie, ACoL, CutMix and ADL, do not have better results than CAM.
These methods have claimed to produce better localization maps, and it turns out that not all of them have succeeded in this direct measurement.
We further apply our SEM model to enhance the extracted localization maps, and it successfully increases the AP scores of four baseline methods.
We notice that SEM decreases the AP scores on ACoL and ADL, which is also caused by high-level erasing operation as we have mentioned.
In Table~\ref{tab:comp-backbones}, we compare the usefulness of SEM on different backbone networks regarding the AP scores.
We find that SEM can consistently make obvious improvements using every backbone network.
InceptionV3~\cite{szegedy2016rethinking} achieves the best scores among the methods without the SEM module, obtaining the AP scores of 70.68\% and 71.06\% on the validation and test sets, respectively.
The proposed SEM method can further increase the AP scores by 2.10\% and 1.48\% and finally obtain the state-of-the-art AP of 72.78\% and 72.54\% on both splits.
SEM can also make consistent improvement on the CUB dataset using various backbone networks, and the improvement of SEM on CUB is even more significant.
In Table~\ref{tab:comp-cub-nets-sal}, the most dramatic increase is made on VGG16, where the AP scores rise by 17.12\% from 59.17\% to 76.29\% via merely using SEM.
SEM can finally boost the AP scores to 77.46\% after rising by 8.56\% when using ResNet50 as the backbone network.
\begin{figure*}[thp]
  \centering
  \includegraphics[width=0.98\textwidth]{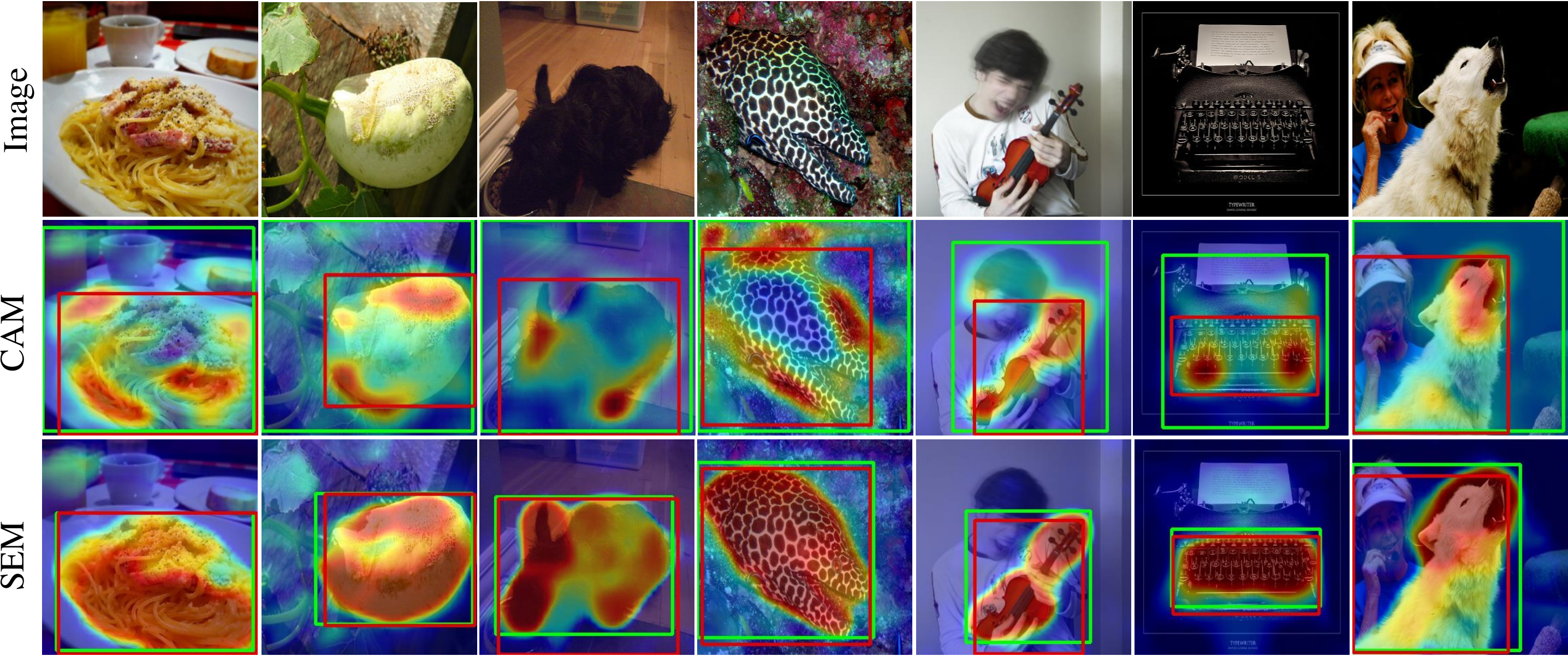}
  \caption{
  Visualization of the localization maps and the predicted bounding boxes with SEM and CAM~\cite{zhou2015cnnlocalization}.
  The ground-truth boxes are in \textcolor{red}{red}, and the predicted boxes are in \textcolor{green}{green}.
  The object regions are more intact and have higher brightness in the maps produced by SEM. 
  }\label{fig-comp-box}
  \vspace{-15pt}
\end{figure*}

Figure~\ref{fig:curves-baselines-c} and ~\ref{fig:curves-baselines-d} compare the Precision-Recall curves between SEM and CAM for extracting localization maps based on recent methods.
We observe that localization maps extracted by SEM generally have higher precision when the recall rate is relatively large ($>0.6$).
The explanation for this is that the higher recall rates are corresponding to the higher thresholds for splitting foreground and background during testing.
Therefore, localization maps extracted by SEM have higher precision of foreground pixels over the higher brightness regions, which is highly consistent with the observation from the proposed IoU-Threshold curves and the visualizations of localization maps.
We can also observe such situations in Figure~\ref{fig:curves-backbones-c} and~\ref{fig:curves-baselines-d}, where various backbone networks are implemented for testing robustness of SEM in different  circumstances.
In Figure~\ref{fig:cub-backbones-b}, such situations are even more obvious on various backbone settings when testing on the CUB dataset.
However, the proposed IoU-Threshold curves are more convenient and easier to observe different properties of localization maps.

\noindent \textbf{Localization Accuracy}
Following the common practice of previous approaches, we also evaluate the proposed SEM using the $indirect$ evaluation metric,~\ie, calculating the percentage of post-inferred bounding boxes that have larger than 50\% IoU with at least one of the ground-truth boxes.
Table~\ref{tab-comp-ilsvrc} compares SEM with various baseline methods. 
We implement SEM on different backbone networks and various recent WSOL methods,~\ie, HaS~\cite{singh2017hide}, CutMix~\cite{yun2019cutmix}, CAM~\cite{zhou2015cnnlocalization} and SPG~\cite{zhang2018self}.
It is clear that our SEM overtakes all of the recent baseline algorithms on different backbone networks under both Top-1 and Gt-known criteria.
We also report the classification error as a reference.
Particularly, ResNet101-SEM achieves the best Top-1 localization accuracy of $54.88\%$, surpassing the other methods and achieving a new state-of-the-art accuracy. 
For the Gt-known localization accuracy,  SEM~\textsubscript{SPG} based on InceptionV3 achieves the best of $69.26\%$ among all the methods, surpassing the current best-performing method,~\ie, HaS, by 3.6\%.
It is also notable that localization accuracy of bounding boxes increases on all the four baselines via simply using SEM to extract localization maps.
The largest increase of Top-1 localization accuracy is acquired by using SEM on CAM, raising by 2.84\% from 50.20\% to 53.04\%.
The corresponding Gt-known localization accuracy also significantly increases by 3.99\% from 65.05\% to 69.04\%.
We also compare the localization accuracy on CUB shown in Table~\ref{tab:comp-cub} with InceptionV3 as the backbone network.
The proposed SEM method surpasses the ADL~\cite{Choe_2019_CVPR} method by a large margin of $8.53\%$, obtaining the Top-1 localization accuracy of $61.57\%$ on CUB.

\noindent \textbf{Visualization}
Figure~\ref{fig-comp-box} compares the localization maps and the post-inferred bounding boxes of various objects on ILSVRC.
We can easily and clearly observe that our SEM overtakes its counterpart,~\ie, CAM, in both visualization effects and predicted bounding boxes.
As we have found in some previous literature~\cite{zhang2018adversarial,zhang2018self,singh2017hide}, CAM is only able to highlight some discriminative and sparse regions of the target objects.
By contrast, localization maps produced by SEM can accurately highlight entire target regions with high confidence.
We show more examples of localization maps on CUB in 
the supplementary material
where the proposed SEM method consistently outperforms the typical way of extracting localization maps,~\ie, CAM.
The localization maps from SEM can generally depict the border information of target objects, which paves the way for us to explore more challenging applications,~\ie, predicting object boundary, in a weakly supervised manner.
As we have discussed in Section~\ref{subsec-metric}, the IoU-Threshold curves and Peak-T values can describe the visualization effects of localization maps in a detailed and quantitative way.
In all the IoU-Threshold curves,~\ie, Figure~\ref{fig:curves-baselines-a},~\ref{fig:curves-baselines-b},~\ref{fig:curves-backbones-a},~\ref{fig:curves-backbones-b} and~\ref{fig:cub-backbones-a}, the peak points drift towards the upper-right corners after using SEM for extracting localization maps.
The meanings of this drifting are 1) more of the target object regions are discovered; 2) brightness of the object regions in localization maps is increased.
The Peak-T values can also be quantitatively compared as shown in Table~\ref{tab:comp-baselines},~\ref{tab:comp-backbones} and~\ref{tab:comp-cub-nets-sal} where the Peak-T values of SEM can consistently outperform the CAM method.

\noindent \textbf{Discussion}
From the above statistics and illustrations, we realize that the recent algorithms~\cite{zhang2018adversarial,zhang2018self,singh2017hide,Choe_2019_CVPR,yun2019cutmix} based on CAM~\cite{zhou2015cnnlocalization} actually have not significantly improved the quality of localization maps.
According to the proposed direct pixel-wise measurement, the quality value of localization maps has only achieved marginal improvements both in accuracy and visualization effects.  
By contrast, SEM shows its superiority for the localization map enhancement.
It makes a significant improvement over various baselines and backbone networks in both accuracy and visualization brightness.
It achieves the state-of-the-art performance on both ILSVRC and CUB benchmarks.
We mainly attribute the success of SEM to 1) the seeds obtained in the first-stage maps reliably lie in the object regions; 2) their features are close to the rest object parts while distant to the background pixels.
For a good performance of SEM, we may not like to implement it on ACoL~\cite{zhang2018adversarial} and ADL~\cite{Choe_2019_CVPR}. 
They erase object regions and force networks to minimize the costs, which brings background noises to high-level feature maps.
The enhanced localization maps could thereby include background regions, and the accuracy of localization maps will decrease.

\begin{table}[t]\setlength{\tabcolsep}{4pt}
  \centering
  \caption{Comparison of the inferred bounding boxes in Gt-known error rates using different number of object seeds. K is the number of object seeds.}\label{tab:ablation_k}
  \vspace{-10pt}
  \small
  \begin{tabular}{ccccccc}
    \toprule[0.2em]
     K & 1 & 20 & 40 & 60 & 80 & 100 \\
    \toprule[0.2em]
    Gt-known Acc & 66.43 & 68.03 & 68.75 & \textbf{69.04} & 68.92 & 68.24 \\ 
    Peak-IoU & 55.59 & 55.99 & \textbf{56.05} & 55.95 & 55.72 & 55.41 \\ 
    \toprule[0.1em]
  \end{tabular}
  \vspace{-15pt}
\end{table}
\vspace{-10pt}
\subsection{Ablation Study for the Number of Seeds}
\textbf{K} object pixels are utilized as seeds to calculate the similarity with the rest pixels for producing the similarity maps.
We expect the K seeds to lie in the target object regions instead of the background areas.
We employ InceptionV3 as the backbone network to study the impact of K.
Table~\ref{tab:ablation_k} compares the Gt-known localization accuracy of the inferred bounding boxes  and the proposed Peak-IoU scores with K changing from 1 to 100.
The Gt-known accuracy increases to $69.04\%$ when K is $60$, which achieves the state-of-the-art record in the weakly supervised localization task. 
When K is getting either larger or smaller, the localization maps will become worse.
The Gt-known performance decreases to $66.43\%$ and $68.24\%$ when K changes to $1$ and $100$, respectively.
The reasons behind this phenomena are that: 
1) When K is too small, the seed pixels are not diverse enough to retrieve all the object pixels;
2) When K is too large, the background pixels might be included, especially if the target objects are too small. Besides, we notice that the best Peak-IoU is achieved when K is chosen as 40. The inconsistency between Peak-IoU and Gt-known accuracy demonstrates that better localization maps cannot guarantee better performance under the \emph{indirect} evaluation metric, which can further prove the necessity of introducing the \emph{direct} evaluation metric.


\vspace{-10pt}
\subsection{Object Boundary Evaluation}
\begin{table}[t]\setlength{\tabcolsep}{12pt}
  \centering
  \caption{Comparison of the predicted object boundaries. 
  SOBD~\cite{uijlings2015situational} is learned and tested on subsets of ILSVRC in fully supervised manner.
  The SEM based methods are obtained with only image-level labels as supervision.
  }
  \vspace{-10pt}
  \begin{tabular}{lccc}
    \toprule[0.2em]
      Method & ODS & OIS & AP \\
    \toprule[0.2em]
     SOBD-class specific~\cite{uijlings2015situational} & - & - & 28.9 \\
     SOBD-subclass specific~\cite{uijlings2015situational} & - & - & 29.6 \\
     SOBD-class agnostic~\cite{uijlings2015situational} & - & - & 29.5 \\
     \hline
     \rowcolor[gray]{0.9}
     CAM-HNS & 34.2 & 33.7 &  25.5 \\
     \hline
     \rowcolor[gray]{0.9}
     SEM-Vanilla & 34.3 &  33.9 &  17.6 \\
     \rowcolor[gray]{0.9}
     SEM-HNS & 39.7 & 37.6 &  31.9 \\
     \rowcolor[gray]{0.9}
     SEM-HNS~\textsubscript{CutMix} & 40.6 & 38.2 &  33.1 \\
     \rowcolor[gray]{0.9}
     SEM-HNS~\textsubscript{HaS} & \textbf{40.8} & \textbf{38.3} &  \textbf{33.3} \\
    \toprule[0.1em]
  \end{tabular}
  \label{tab:comp-edge}
  \vspace{-15pt}
\end{table}
Based on the localization maps of SEM, we further explore to predict object boundaries in a weakly supervised manner on a very large-scale dataset.
To examine the quality of the inferred object boundaries, we follow the common practice of fully supervised edge detection approaches~\cite{he2019bi} to conduct the evaluation. 
Particularly, we apply a standard non-maximal suppression (NMS) technique to the predicted edge maps to obtain thinned edges, and then report the performance with the most commonly used metrics,~\ie, Average Precision (AP), Optimal Dataset Scale (ODS) and Optimal Image Scale (OIS). 
We infer the ground-truth boundaries from the annotated masks of the validation set and employ the popular edge toolbox in~\cite{DollarICCV13edges,ZitnickECCV14edgeBoxes} for evaluation. The maximum tolerance allowed for correct matches between edge predictions, and the ground-truth edges is set to 0.0075.

\begin{figure}[t]
  \centering
  \includegraphics[width=0.48\textwidth]{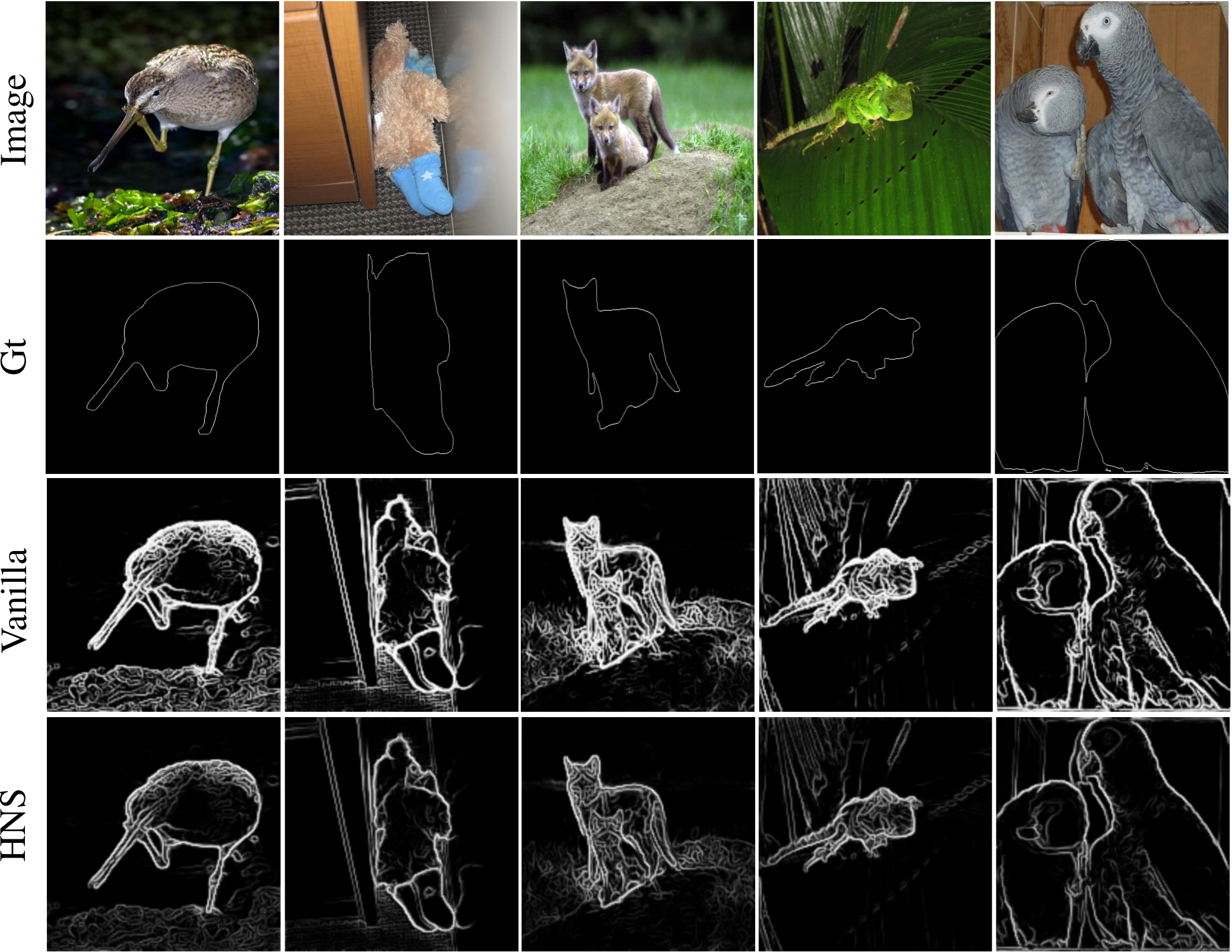}
  \caption{
  The predict the object boundaries using only image categories as supervision.
  Our HNS loss is able to effectively suppress unimportant noises.
  }\label{fig-comp-edge}
  \vspace{-15pt}
\end{figure}

We choose InceptionV3~\cite{szegedy2016rethinking} as the backbone network.
In Table~\ref{tab:comp-edge}, we compare different methods in predicting object boundaries on ILSVRC.
First, we compare SEM-Vanilla with SEM-HNS where the former one is learned using the vanilla loss function in Eq.~\eqref{eq_vanilla_edge}, and the latter one is trained by adding the proposed HNS item.
Surprisingly, the proposed HNS loss function can dramatically increase the AP score by 14.3\% from 17.6\% to 31.9\%. 
By adopting useful data augmentation techniques in HaS~\cite{singh2017hide} and CutMix~\cite{yun2019cutmix}, the AP scores can be further enhanced to 33.3\% and 33.1\%, respectively.
Next, we verify the influence of localization maps between the proposed SEM and CAM~\cite{zhou2015cnnlocalization}. 
By replacing localization maps from SEM to CAM during the learning object boundaries stage, the AP score decreases from 31.9\% to 25.5\%.
This result indicates that our proposed SEM can discover more details in the produced localization maps than CAM.
As a reference, we try our best and find only one method,~\ie, SOBD~\cite{uijlings2015situational}, which aims at predicting object boundaries on such a large scale dataset.
Different from our method, SOBD adopts a fully supervised approach to learning multiple situational detectors.
It is surprising to see that our weakly-supervised edge detector significantly outperforms SOBD by more than 3.8\% in AP.
Figure~\ref{fig-comp-edge} compares the predicted object boundaries between the vanilla loss function and our modified HNS function.
Caused by the lack of ground-truth boundaries, our edges include noises within the object regions.
After applying the proposed hard negative depression loss function, we can reduce the background edges and weaken some unwanted edges.

\vspace{-10pt}
\section{Conclusion}
In this paper, we make three contributions. 
We firstly analyze the deficiencies of the current indirect metric and propose to apply a more delicate and direct approach for evaluating localization maps. 
We annotate images on the ILSVRC validation set to fulfill the direct metric.
Then, we propose a two-stage method for obtaining better localization maps, which can accurately cover target objects.
The proposed SEM method outperforms all existing methods with various backbone networks.
Based on SEM, we further explore the approach for generating quality object boundaries with only image-level labels as supervision.
The proposed SEM-HNS method can accurately predict object boundary pixels and also outperform baseline methods in Average Precision.
SEM-HNS applies the proposed Hard-Negative-Suppression loss to eliminate undesired edges in the background.

\vspace{-10pt}
\bibliographystyle{IEEEtran}
\small
\bibliography{IEEEabrv,body/ref}

%
%






\end{document}


%
\title{Supplementary Material}
\markboth{IEEE TRANSACTIONS ON PATTERN ANALYSIS AND MACHINE INTELLIGENCE}
{Shell \MakeLowercase{\textit{et al.}}: Bare Demo of IEEEtran.cls for Computer Society Journals}
\maketitle

\IEEEdisplaynontitleabstractindextext

%
\IEEEpeerreviewmaketitle

\appendices 
\section{Mask Annotation on ILSVRC}~\label{supp_annotation}
\begin{figure}[bhp]
  \centering
  \includegraphics[width=0.48\textwidth]{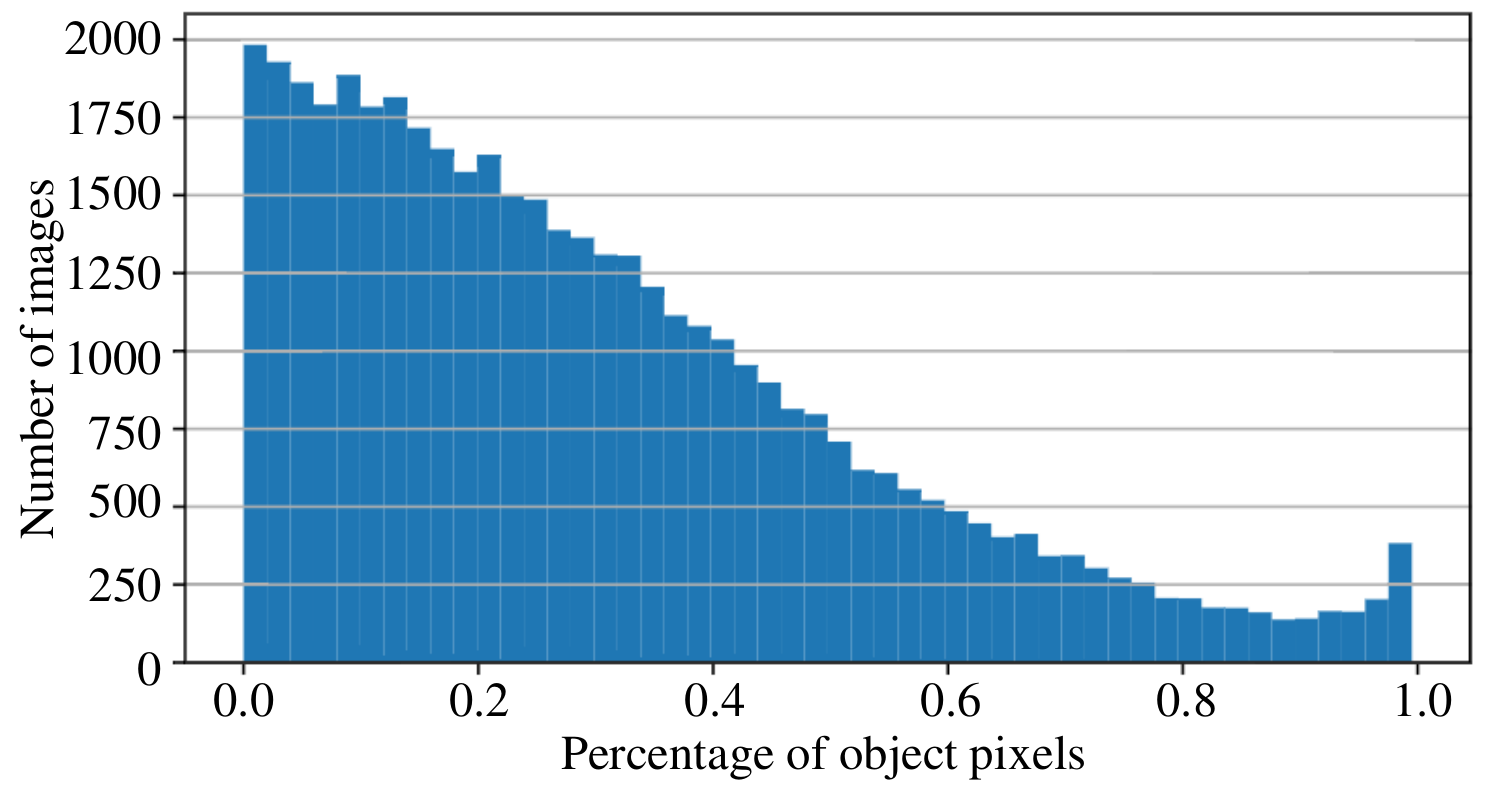}
  \caption{Histogram of object area rates in images on ILSVRC.
  }\label{fig:fg_rate}
\end{figure}

\begin{figure*}[!thp]
  \centering
  \includegraphics[width=0.98\textwidth]{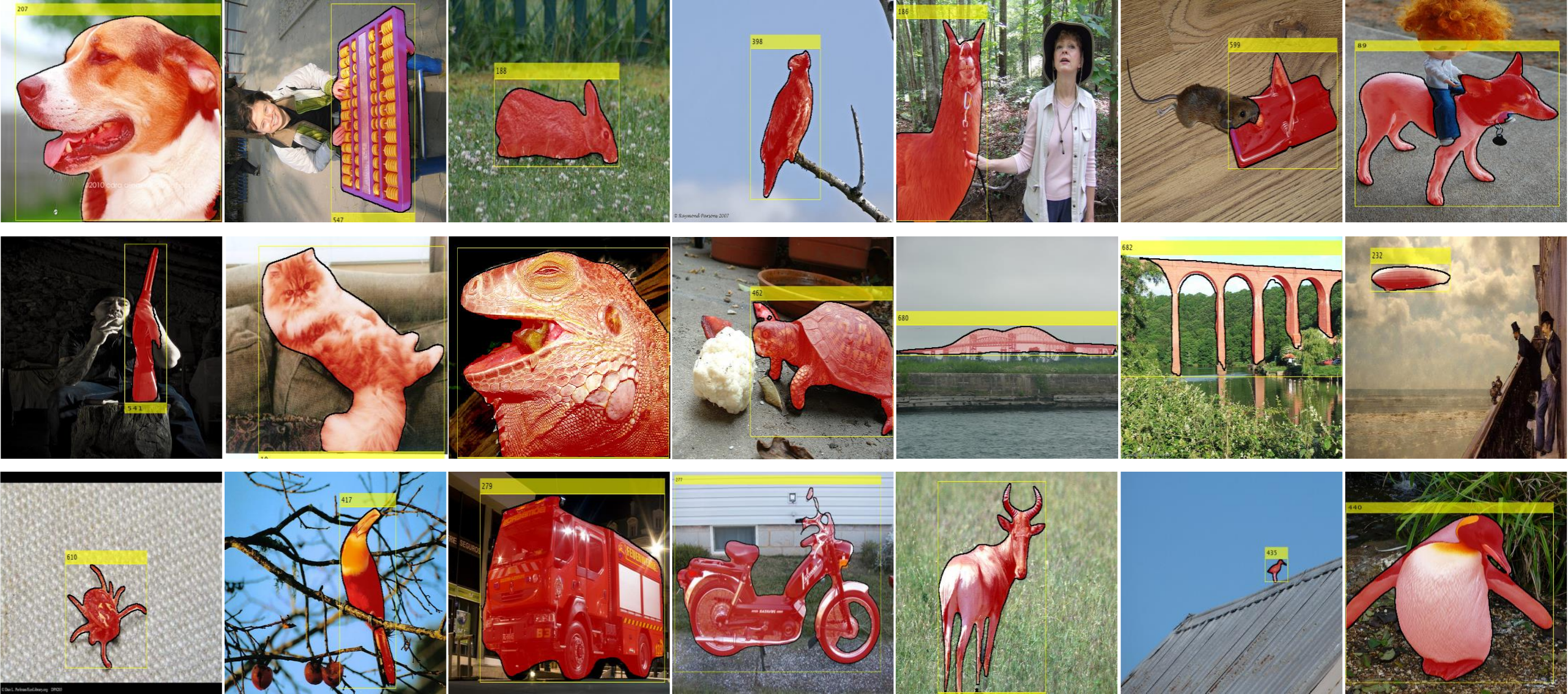}
  \caption{
  Illustration of our annotated object masks on the ILSVRC validation set. 
  Numbers indicate the category IDs in ILSVRC~\cite{ILSVRC15}
  }\label{fig:gt_mask}
\end{figure*}

In order to fulfill the purpose of directly evaluating localization maps with high precision and reliability, we annotate pixel-level masks of objects on the ILSVRC~\cite{2009-imagenet} validation dataset which is the most acknowledged dataset for studying localization maps. 
We first manually exclude 5,729 images that have ambiguous object pixels and annotate the left 44,271 images.
We then divide the annotated masks into two groups,~\ie, the validation and test group.
The validation group contains 23,151 images, while the test group contains 21,120 images.
Figure~\ref{fig:gt_mask} depicts some images with the labeled pixel-level masks.
The target objects are chosen according to the image-level labels and bounding boxes provided on the ILSVRC CLS-LOC dataset.
If multiple objects appear in a given image,~\eg, the man holding a rifle, we only label the object corresponding to the image-level label,~\eg, rifle.
We choose such an annotation strategy because the annotated masks should be consistent with the category labels for training classification networks.
Figure~\ref{fig:fg_rate} shows the histogram of the area of target objects of all the 44,271 images.
It is obvious that objects account for less than half of the area in most images.
The scale of Region of Interests (RoIs) changes greatly, as the smallest objects only have about 10 pixels while the largest objects contain about 1M pixels.
Also, we notice that there are about 400 images where target objects occupy nearly the entire images.

\begin{figure*}[!thp]
  \centering
  \includegraphics[width=0.98\textwidth]{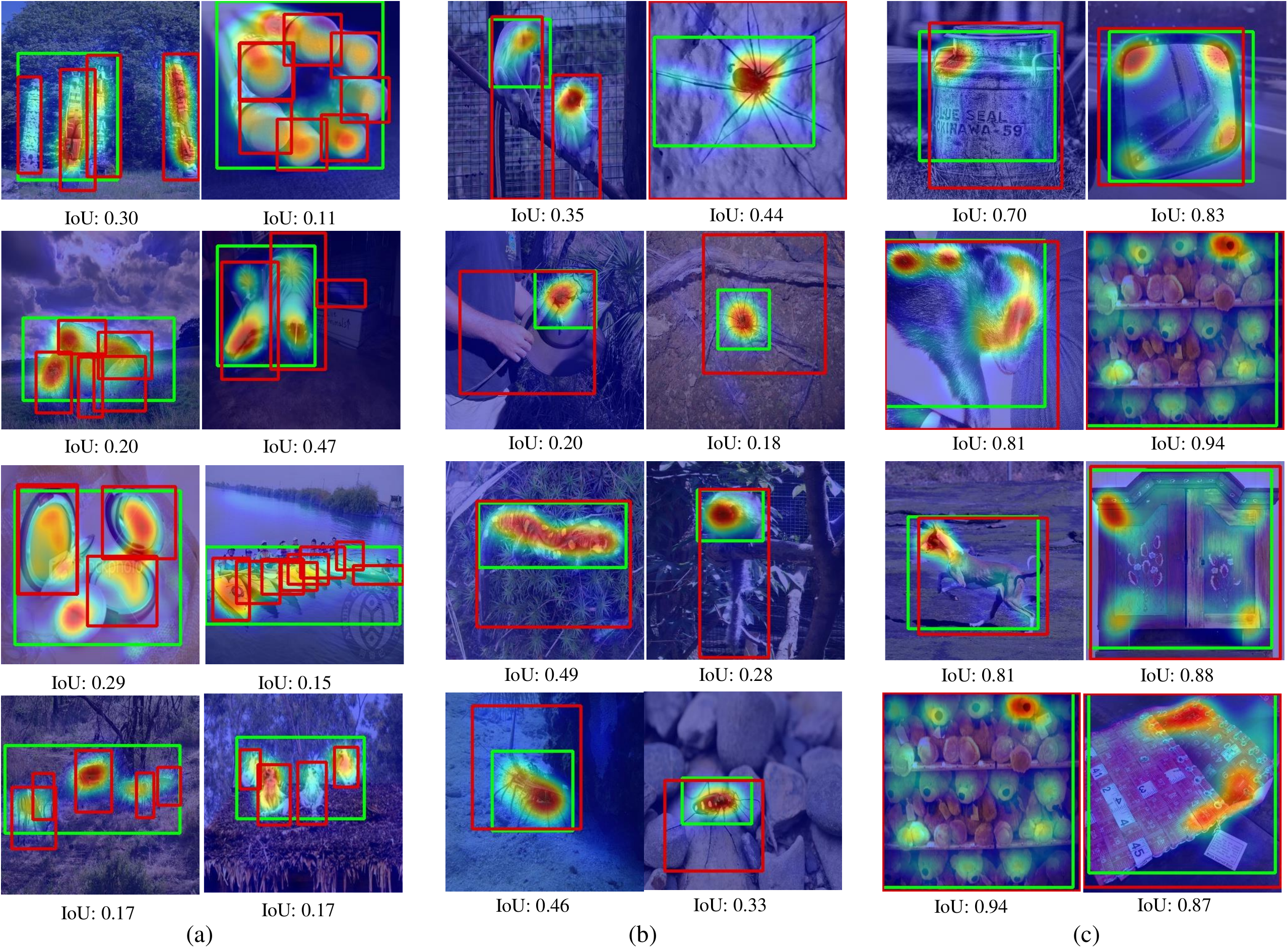}
  \caption{
  The current metric treats an image as ture positive if the post-inferred bounding boxes have over 50\% IoU with at least one of the ground-truth boxes. 
  Here, we show three scenarios where the current indirect evaluation metric fails to measure localization maps: 
  (a) Localization maps accurately highlight target objects, while the metric considers them as false positive; (b) main parts of the target objects are highlighted,  while the metric gives them zero credit; (c) although the predicted bounding boxes are accurate, the localization maps fail to highlight important areas of the target objects. 
  }\label{sup:three_cases}
\end{figure*}
\begin{figure*}[!thp]
  \centering
  \includegraphics[width=1.0\textwidth]{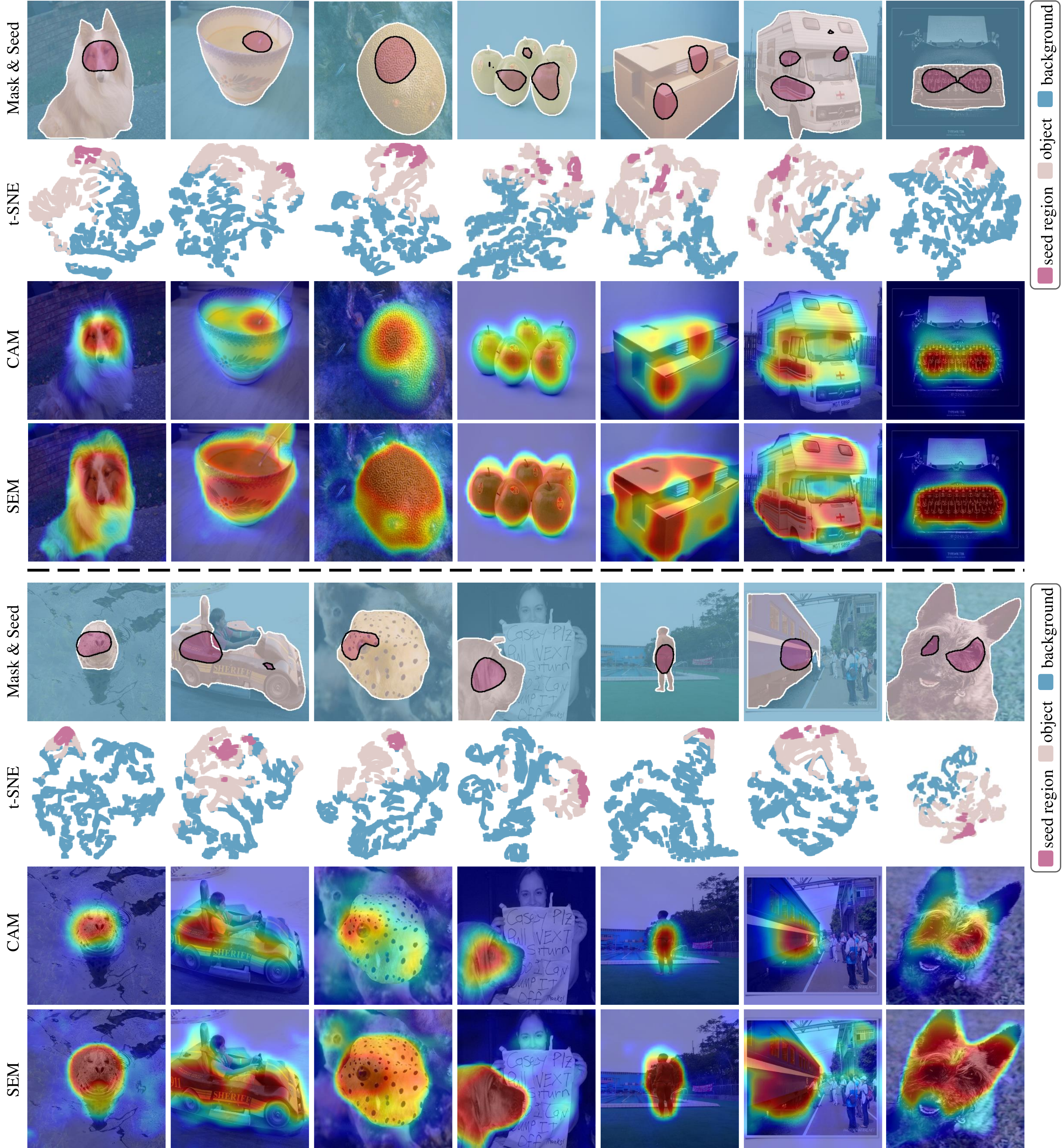}
  \caption{
   t-SNE~\cite{van2014accelerating} of high-level features.
   Motivation illustration and visualization comparison of the proposed SEM approach.
   We illustrate the seed regions using the pixels with top 70\% scores chosen in SPG~\cite{zhang2018self}.
   The top two rows show the high-level features of seed regions are closer to the rest object regions than background areas. 
   We implement similarities between pixels to enhance localization maps.
   The bottom two rows compare the proposed SEM and CAM~\cite{zhou2015cnnlocalization}, and SEM achieves better performance in accurately highlighting target objects.
  }\label{sup:three_cases}
\end{figure*}

\section{SEM with different feature maps}~\label{supp_feat_pos}
Feature maps of different depths are commonly acknowledged to have different semantic information.
Features from low-level layers tend to involve more edge and texture information, while features from high-level layers are more abstract to express rough object locations.
We employ feature maps of different depths to study the changes in localization abilities.
In Table~\ref{tab:ablation_featpos},  we compare the Gt-known localization accuracy and Peak-IoU with feature maps of different depths.
$feat3$ and $feat4$ are the output feature maps of Block3 and Block4. 
$feat5$ is the output feature maps of the penultimate convolutional layer.
Please refer to the  \href{https://github.com/xiaomengyc/SEM}{released code} for more details.
Gt-known localization accuracy is the best when using the feature maps of the penultimate convolutional layer (\textit{labeled as $feat5$ in Table~\ref{tab:ablation_featpos}}), achieving the highest accuracy of 69.04\%.
The localization ability is dramatically getting worse when applying feature maps of lower levels.
We also study the Gt-known localization accuracy of the combination of $feat4$ and $feat5$ by concatenation. 
By such a method of feature map concatenation, the localization accuracy decreases to $68.03\%$, which is worse than the result of only using $feat5$.
In terms of the proposed direct metric,~\ie, Peak-IoU, it achieves 56.06\% when only using the features from the fifth block.
Peak-IoU can be slightly increased to 56.09\% when concatenating \emph{feat3} and \emph{feat5}.
In our experiments,  we only use \emph{feat5} to obtain localization maps using the proposed SEM approach. 

\begin{table}[t]\setlength{\tabcolsep}{6pt}
  \caption{
  Comparison of the inferred bounding boxes in Gt-known accuracy using different feature maps. 
  \textit{featx}  denote the feature maps of $x_{th}$ block.
  feat5 is extracted from the penultimate convolution layer in the $5_{th}$ block.
  }\label{tab:ablation_featpos}
  \centering
  \small
  \begin{tabular}{ccccc}
    \toprule[0.2em]
     feat3 & feat4 & feat5 & Gt. known Acc & Peak-IoU \\
    \toprule[0.2em]
       \checkmark &  &   & 42.55 & 40.98 \\
     & \checkmark &   & 58.56 & 52.41 \\
     &  &  \checkmark & \textbf{69.04} & 56.05\\
     & \checkmark &  \checkmark & 68.03 & \textbf{56.09} \\
    
    \toprule[0.1em]
  \end{tabular}
\end{table}

\begin{figure*}[t]
  \centering
  \includegraphics[width=0.98\textwidth]{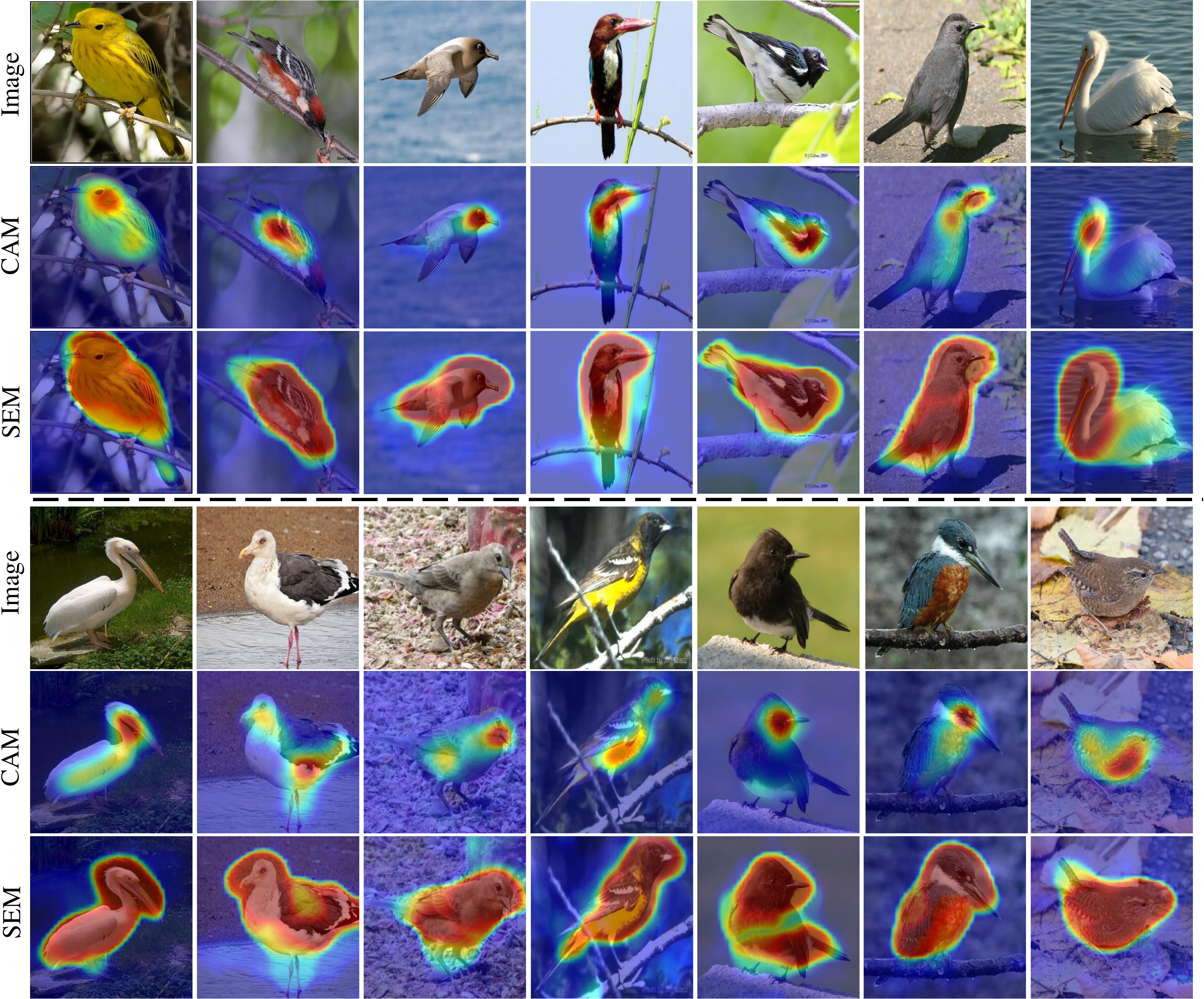}
  \caption{
  Visualization of the localization maps produced by CAM~\cite{zhou2015cnnlocalization} and SEM on CUB~\cite{WahCUB_200_2011}.
  CAM only highlights the sparse discriminative regions, while SEM can highlight the entire object regions and present more distinctions between the objects and background.
  }\label{fig-cub-box}
  \vspace{-20pt}
\end{figure*}

\bibliographystyle{IEEEtran}
\small
\bibliography{IEEEabrv,body/ref}
%
%







